\documentclass[preprint,12pt]{elsarticle}
\usepackage[left=2cm,right=2cm,top=2cm,bottom=2cm]{geometry} % Tùy chỉnh lề nếu cần
\usepackage{amsmath}
\usepackage{hyperref}
\hypersetup{
    colorlinks=true, % Set to true to enable colored links
    citecolor=green,  % Set citation color
    linkcolor=red,  % Set the color of internal links (sections, pages, etc.)
    urlcolor=blue    % Set the color of web external hyperlinks
}
\usepackage{subcaption}
\usepackage{lipsum}
\usepackage{amsmath,amssymb,amsfonts}
\usepackage{mismath}
\usepackage{xcolor}
\usepackage[linesnumbered,ruled,vlined]{algorithm2e}
\usepackage{graphicx}
\usepackage{algpseudocode}
\usepackage{booktabs,caption}
\usepackage{textcomp}
\usepackage{tabularx}
\usepackage{xcolor}
\usepackage{amssymb}% http://ctan.org/pkg/amssymb
\usepackage{pifont}% http://ctan.org/pkg/pifont
\newcommand{\xmark}{\ding{56}}%
\usepackage{colortbl}

\definecolor{gray1}{gray}{0.9}
\definecolor{gray2}{gray}{0.8}
\definecolor{gray3}{gray}{0.7}
\definecolor{gray3}{gray}{0.92}
\usepackage{arydshln}

\usepackage{multirow}
\usepackage{algpseudocode}% <========== http://ctan.org/pkg/algorithmicx
\usepackage{hyperref} % <===============================================
\usepackage{hyperref}
\usepackage{amsmath,amssymb,amsfonts}
\usepackage{graphicx}
\usepackage[utf8]{inputenc}
\usepackage[english]{babel}
\usepackage{indentfirst}
\usepackage{booktabs}
\usepackage{tabularx}
\usepackage[pscoord]{eso-pic}

\usepackage{amsmath}

\journal{}
\makeatletter
\def\ps@pprintTitle{%
   \let\@oddhead\@empty
   \let\@evenhead\@empty
   \let\@oddfoot\@empty
   \let\@evenfoot\@oddfoot
}
\makeatother

\begin{document}
% \linenumbers

\begin{frontmatter}

\title{\textbf{TwinLiteNet$^+$: An Enhanced Multi-Task Segmentation Model for Autonomous Driving}}

\author[First,Second]{Quang-Huy Che}
\ead{huycq@uit.edu.vn}

\author[First,Second]{Duc-Tri Le}
\ead{21522703@gm.uit.edu.vn}

\author[First,Second]{Minh-Quan Pham}
\ead{quanpm@uit.edu.vn}

\author[First,Second]{Vinh-Tiep Nguyen}
\ead{tiepnv@uit.edu.vn}

\author[First,Second]{Duc-Khai Lam\corref{cor1}}
\ead{khaild@uit.edu.vn}

% \cortext[eq]{First two authors contribute equally.}
\cortext[cor1]{Corresponding author.}

\address[First]{University of Information Technology, Ho Chi Minh City, Vietnam}
\address[Second]{Vietnam National University, Ho Chi Minh City, Vietnam.}
\begin{abstract}

% The quest for robust Person re-identification (Re-ID) systems capable of accurately identifying subjects across diverse scenarios remains a formidable challenge in surveillance and security applications. This study presents a novel methodology that significantly enhances Person Re-Identification (Re-ID) by integrating Uncertainty Feature Fusion (UFFM) with Wise Similarity Aggregation (AMC). Tested on benchmark datasets - Market-1501, DukeMTMC-ReID, and MSMT17 - our approach demonstrates substantial improvements in Rank-1 accuracy and mean Average Precision (mAP). Specifically, UFFM capitalizes on the power of feature synthesis from multiple images to overcome the limitations imposed by the variability of subject appearances across different views. AMC further refines the process by intelligently aggregating similarity metrics, thereby enhancing the system's ability to discern subtle but critical differences between subjects. The empirical results affirm the superiority of our method over existing approaches, achieving new performance benchmarks across all evaluated datasets. Code is available on Github.

Semantic segmentation is a fundamental perception task in autonomous driving, particularly for identifying drivable areas and lane markings to enable safe navigation. However, most state-of-the-art (SOTA) models are computationally intensive and unsuitable for real-time deployment on resource-constrained embedded devices. In this paper, we introduce TwinLiteNet$^+$, an enhanced multi-task segmentation model designed for real-time drivable area and lane segmentation with high efficiency. TwinLiteNet$^+$ employs a hybrid encoder architecture that integrates stride-based dilated convolutions and depthwise separable dilated convolutions, balancing representational capacity and computational cost. To improve task-specific decoding, we propose two lightweight upsampling modules-Upper Convolution Block (UCB) and Upper Simple Block (USB)-alongside a Partial Class Activation Attention (PCAA) mechanism that enhances segmentation precision. The model is available in four configurations, ranging from the ultra-compact TwinLiteNet$^+_{\text{Nano}}$ (34K parameters) to the high-performance TwinLiteNet$^+_{\text{Large}}$ (1.94M parameters). On the BDD100K \cite{bdd} dataset, TwinLiteNet$^+_{\text{Large}}$ achieves 92.9\% mIoU for drivable area segmentation and 34.2\% IoU for lane segmentation-surpassing existing state-of-the-art models while requiring 11× fewer floating-point operations (FLOPs) for computation. Extensive evaluations on embedded devices demonstrate superior inference speed, quantization robustness (INT8/FP16), and energy efficiency, validating TwinLiteNet$^+$ as a compelling solution for real-world autonomous driving systems. Code is available at \url{https://github.com/chequanghuy/TwinLiteNetPlus}.

% Semantic segmentation is crucial for autonomous driving, particularly for the tasks of Drivable Area and Lane Segmentation, ensuring safety and navigation. To address the high computational costs of current state-of-the-art (SOTA) models, this paper introduces TwinLiteNetPlus (TwinLiteNet$^+$), a model capable of balancing efficiency and accuracy. TwinLiteNet$^+$ incorporates standard and depthwise separable dilated convolutions, reducing complexity while maintaining high accuracy. It is available in four configurations, from the robust 1.94 million-parameter TwinLiteNet$^+_{\text{Large}}$ to the ultra-lightweight 34K-parameter TwinLiteNet$^+_{\text{Nano}}$. Notably, TwinLiteNet$^+_{\text{Large}}$ attains a 92.9\% mIoU (mean Intersection over Union) for Drivable Area Segmentation and a 34.2\% IoU (Intersection over Union) for Lane Segmentation. These results achieve remarkable performance, surpassing current state-of-the-art models while only requiring 11 times fewer Floating Point Operations (FLOPs) for computation. Rigorously evaluated on various embedded devices, TwinLiteNet$^+$ demonstrates promising latency and power efficiency, underscoring its potential for real-world autonomous vehicle applications.

\end{abstract}


\begin{keyword}
Self-Driving Car \sep Lane segmentation \sep Drivable segmentation \sep BDD100K \sep Light-weight model \sep Embedded devices
\end{keyword}

\end{frontmatter}

\section{Introduction}

The emergence of deep learning methodologies has driven significant growth in the field of autonomous vehicles, where visual perception plays a central role. Among various sensors, cameras remain the most cost-effective and offer high-resolution RGB data for scene understanding. Compared to LIDAR and Radar, which provide depth but are costly and color-insensitive, cameras offer a practical trade-off between cost computation and semantic detail. As a result, camera-based perception has become a dominant focus in the development of real-time driving systems, particularly those powered by convolutional neural networks.

\begin{figure}[!t]
    \centering
        \includegraphics[width=0.9\linewidth]{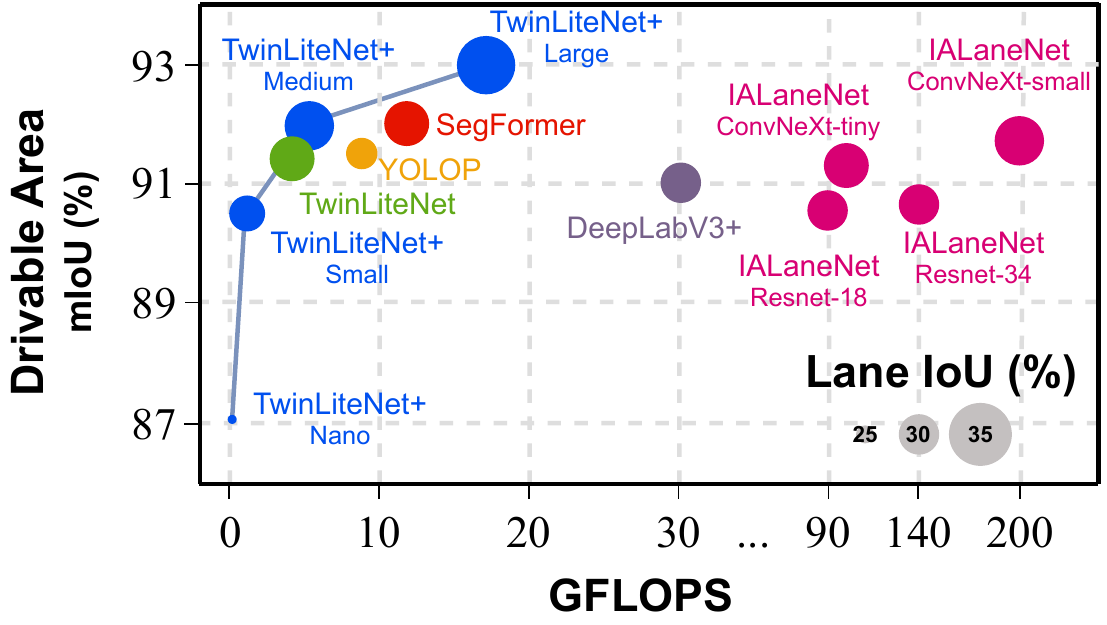}
        \caption{
        Comparison of evaluation metrics mIoU (\%) (for Drivable Area Segmentation) - IoU (\%) (for Lane Segmentation) - GFLOPs of various models on the BDD100K dataset. The x-axis represents the number of GFLOPs of the model, the y-axis represents the mIoU (\%) for the Drivable Area Segmentation task, and the radius of the circle indicates the IoU (\%) for the Lane Segmentation task (the larger the radius, the higher the accuracy).}
        \label{model_comparison}
\end{figure}

In Advanced Driver Assistance Systems (ADAS) \cite{adas1,adas2,adas3}, camera-based panoptic driving perception serves as a critical component. By capturing comprehensive semantic information about the surrounding environment from images, vehicles can make optimal driving decisions. Prior studies \cite{enet,enetsad,pspnet,multinet,scnn,interactive,BILane,twin} have highlighted that segmentation plays a fundamental role in enabling ADAS to react promptly and accurately. Specifically, (1) \textit{Drivable area segmentation} identifies regions where the vehicle can safely operate, thereby supporting precise control decisions. (2) \textit{Lane segmentation} provides detailed insights into lane structure and positioning, assisting the vehicle in maintaining or changing lanes safely. While models are trained on large-scale datasets \cite{ss1,ss2,deeplab,segformer,espnet,espnetv2} such as Cityscapes \cite{cityscapes} offer detailed scene-level semantic labels, their granularity is often less relevant for immediate control decisions. In contrast, directly segmenting drivable regions and lanes provides more targeted and operationally helpful outputs for autonomous navigation.

Recent research has increasingly shifted from single-task models \cite{enet,enetsad,pspnet,multinet,scnn} to multi-task learning models \cite{interactive,BILane,twin}, which aim to address multiple segmentation tasks concurrently. These models exploit shared encoder representations to reduce computational redundancy while maintaining task-specific decoder branches to preserve accuracy. However, achieving high performance often needs architectural complexity models, which may be impossible in real-world deployment. Despite being designed for real-time applications, many existing multi-task approaches \cite{interactive,hybridnets,yolopx,yolopv3} are evaluated primarily on high-end GPUs, ignoring the strict constraints posed by embedded systems-such as limited processing power, energy efficiency, and memory capacity.

In this work, we introduce TwinLiteNet$^+$, a model tailored for real-time operation with optimal power consumption and hardware resources, proficient in simultaneously segmenting lanes and drivable areas. This model demonstrates competitive precision compared to models with similar tasks. There are four principal contributions of our research:

\begin{itemize}
    \item We introduce TwinLiteNet$^+$, a lightweight and computation-efficient multi-task model designed for drivable area and lane segmentation. The encoder leverages a hybrid design of stride-based dilated convolutions and depthwise dilated separable convolutions to balance receptive field size and computational cost. In the decoder, we introduce Upper Convolution Block (UCB) and Upper Simple Block (USB) modules, which enhance upsampling efficiency without introducing heavy computation, making the architecture suitable for real-time inference.

    \item TwinLiteNet$^+$ is implemented in four configurations-Nano, Small, Medium, and Large-ranging from just 0.03M parameters / 0.57 GFLOPs (Nano) to 1.94M parameters / 17.58 GFLOPs (Large). This range of configurations enables flexible deployment across devices with varying computational capabilities, from ultra-low-power embedded systems to higher-end edge devices.

    \item Experimental results on the BDD100K dataset show that TwinLiteNet$^+_{\text{Large}}$ achieves 92.9\% mIoU for drivable area segmentation and 34.2\% IoU for lane segmentation, outperforming state-of-the-art lightweight models while maintaining low model complexity.

    \item We deploy TwinLiteNet$^+$ on real-world embedded devices such as NVIDIA Jetson TX2 and Jetson Xavier, demonstrating real-time inference performance with low latency and efficient energy consumption. Furthermore, we show that the model is highly robust under INT8 and FP16 quantization settings.
\end{itemize}

% (1) We present a lightweight Convolutional Neural Network (CNN) model, optimized for low computational cost, consisting of a singular encoder block and dual decoder blocks for drivable area and lane segmentation respectively. (2) The model is available in four distinct configurations, each optimizing the trade-off between accuracy and computational efficiency. (3) To validate our model's applicability in real-world conditions, we deploy and assess its performance in terms of speed and energy efficiency on a range of embedded devices, including the Jetson Xavier and Jetson TX2. 

The remainder of the paper is represented as follows: We evaluate relevant models in Section \ref{related} to grasp the benefits and drawbacks in the tasks of  Drivable Area Segmentation, and Lane Segmentation with Multi-task approaches. The proposed TwinLiteNet$^+$ presents an architecture with methods to boost model performance in Section \ref{Propose}. In Section \ref{result}, we conduct experiments on the BDD100K dataset, and the results show that our TwinLiteNet$^+$ models outperform on latency and power efficiency. In the final Section, we provide some conclusions and future development directions.

\section{Related Work}\label{related}

\subsection{\textbf{Semantic Segmentation for autonomous driving}}
Semantic segmentation is a fundamental subfield in computer vision that aims to assign a class label to each pixel in an input image. Unlike tasks such as image classification or object detection, semantic segmentation provides fine-grained information by precisely identifying the shape and boundaries of each object within a scene. Research in this area has primarily focused on scene understanding \cite{ss1,ss2,deeplab,segformer,espnet,espnetv2,mask2former}, with models typically trained and evaluated on densely annotated datasets such as Cityscapes \cite{cityscapes}. Modern segmentation models are often built upon deep learning architectures and further enhanced through the integration of attention modules \cite{cfam,senet,pcaa,cbam}, which help the model focus on informative regions and improve spatial representation learning. Although these approaches have achieved high performance in modeling complex scene contexts, they are not well suited for real-time applications such as autonomous driving, where computational efficiency are critical. In such scenarios, a more practical approach is to focus on task-relevant functional regions-such as drivable areas and lane segmentation-to enable timely and reliable driving decisions.
\subsubsection{\textbf{Signle-task Approaches}}

Drivable area segmentation enables models to identify navigable regions of the road directly. Several existing approaches \cite{multinet,pspnet} handle the drivable area as a single unified class, encompassing both direct and alternative drivable areas. In contrast, more advanced models \cite{enhanced, shuda, aspp, fast, lightweight, idsmodel} have demonstrated the ability to distinguish between direct and alternative areas when trained on comprehensive datasets such as BDD100K, thereby offering a more understanding of the drivable space. However, segmenting the road into fine-grained components such as individual lanes-particularly differentiating between vertical lanes and parallel lanes (e.g., stop lines, crosswalks)-requires lane detection models. Parashar et al.\cite{scnn} proposed Spatial CNN (SCNN) with a new neural network architecture by generalizing traditional deep layer-by-layer to slice-by-slice convolution in feature mapping, allowing information to propagate across pixels along rows and columns within a layer despite time-consuming. On the other hand, Enet-SAD \cite{enetsad} utilizes a self-attention-guided filter method to assist low-level feature maps in learning from high-level feature maps, or recently, road marking methods \cite{pinet, ultra1, ultra2,QuantLaneNet} has also brought much attention from the community. A novel detector CLRerNet \cite{clrernet} along with a proposed LaneIoU metric are presented to improve the confidence score further.

Although previous research highlighted the initial success of models in the single-task domain, the advantage of leveraging a model to address multiple tasks has led to recent studies demonstrating that a multi-task strategy is more appropriate for real-world applications, particularly when deployed on devices with limited computing capabilities, such as embedded systems. This method allows models to strike a balance between efficiency and complexity while concurrently performing two tasks: segmentation of the driver's area and lane segmentation. Motivated by these advantages, we propose a multi-task architecture with a shared encoder and dual decoders, targeting both drivable area and lane segmentation tasks simultaneously.

\subsection{\textbf{Multi-task Approaches}}

In recent years, multi-task learning has gained considerable attention in autonomous driving, particularly for jointly addressing segmentation-related tasks such as drivable area segmentation, lane segmentation, and scene parsing. By leveraging shared representations among these spatially correlated tasks, multi-task frameworks can improve model generalization while significantly reducing computational overhead-an essential consideration for real-time applications. The availability of large-scale datasets such as BDD100K \cite{bdd} and NYUv2 \cite{NYUv2} has further accelerated research in this direction, enabling the development and benchmarking of various segmentation-based multi-task models. Several studies extend to auxiliary segmentation tasks, combining semantic segmentation with scene classification \cite{vislab, fast} or depth estimation \cite{Jacob_2023_WACV,swinmtl} to enhance environmental understanding. Others have adopted dual-task configurations, such as jointly performing drivable area and lane segmentation \cite{twin, BILane, interactive, yolop, segformer, deeplab, sparse}, which are critical for navigation and path planning.

Among prior multi-task segmentation approaches, some methods \cite{deeplab,segformer} repurpose single-task segmentation backbones and adapt them to multi-task settings. At the same time, \cite{yolop,sparse} are initially designed for joint detection and segmentation, yet provide segmentation-only configurations. However, these models are not optimized explicitly for segmentation synergy, often resulting in limited performance for multi-task segmentation. In contrast, models like IALaneNet \cite{interactive} explicitly focus on inter-task feature fusion using attention mechanisms at the task level. These methods demonstrate improved cooperation between segmentation branches but are limited by complex architectures and high computational costs, making them less suitable for real-time or embedded deployment. To address the limitations of practical deployment, lightweight models such as TwinLiteNet and BILane \cite{twin, BILane} have been proposed. TwinLiteNet, with only 0.4M parameters, is optimized for real-time inference on embedded platforms, though its shallow decoder constrains performance in challenging lane scenarios. BILane \cite{BILane} improves upon this with boundary-aware and interactive attention modules in a compact 1.4M-parameter design. However, its segmentation accuracy remains not as strong compared to TwinLiteNet despite using approximately 3.5 times more parameters and relying on higher input image resolution. These works clearly show a trade-off between segmentation accuracy and inference speed. This motivates the need for a multi-task segmentation model that can balance accuracy and efficiency.

Inspired by TwinLiteNet \cite{twin}, the proposed TwinLiteNet$^+$ introduces a more powerful yet efficient architecture by enhancing both the encoder and decoder. The decoder is improved with additional convolutional layers after transposed convolutions to reconstruct fine-grained features better, while the encoder is designed to optimize computational efficiency without compromising feature extraction capability. Moreover, TwinLiteNet$^+$ comes in four scalable configurations, enabling flexible deployment across a wide range of hardware platforms-from high-performance GPUs to resource-constrained embedded systems.

\subsection{\textbf{Optimizing Deep Learning Deployment on Embedded Devices}} \label{2c}
In recent years, deploying deep learning models on embedded devices has attracted considerable attention within the artificial intelligence community. The computation on these devices is distinguished by its ability to execute processes directly on units that are both cost-effective and task-specific. Despite this advantage, such capabilities are constrained by significant limitations in terms of computational power and storage capacity, presenting substantial challenges for the deployment of deep learning models on these platforms. To address these limitations, several model compression techniques have been proposed, such as 8-bit quantization \cite{quantization}, pruning \cite{pruning}, and knowledge distillation \cite{knowstill}. Among these, quantization is widely favored for its simplicity and hardware-friendliness. It reduces model precision to lower bit widths (e.g., INT8 or FP16), significantly decreasing memory usage and inference latency. Post-training quantization (PTQ) allows models trained in 32-bit floating-point to be quantized without retraining, whereas quantization-aware training (QAT) simulates quantization effects during training and typically yields higher accuracy. In contrast, pruning requires careful exploration of pruning schedules and sensitivity analysis across layers, often requiring iterative retraining to restore accuracy. This process is time-consuming and highly dependent on manual tuning or costly search algorithms, and it may not be generalized well across architectures. Similarly, knowledge distillation introduces a complex teacher-student training pipeline where a compact model (student) is trained to mimic the output distribution of a larger pre-trained model (teacher). This multi-stage training not only adds overhead but also does not offer guarantees about interpretability or alignment with the original model’s decision boundaries.

Howard et al. introduce MobileNet \cite{mobilenet}, a class of efficient neural network architectures tailored for embedded and mobile vision applications, leveraging depthwise separable convolutions to reduce both parameter count and computational complexity significantly. Based upon this foundation, some studies have shown that depthwise dilated separable convolutions outperform standard dilated convolutions in classification tasks, achieving 67.9\% accuracy with only 123M FLOPs-compared to 69.2\% accuracy with 478M FLOPs using standard dilation-highlighting that the latter offers only marginal gains in accuracy at nearly 4× the computational cost. Beyond architectural improvements, numerous model compression and optimization strategies have been proposed for low-resource deployment \cite{shufflenet,ghostnet}. Frameworks such as TensorFlow Lite, TensorRT, NCNN, and MNN \cite{mnn} have played a pivotal role in facilitating deployment by supporting quantization, operator fusion, and backend-specific accelerations.

Despite significant progress, achieving an optimal trade-off between accuracy, latency, and computational efficiency remains a persistent challenge-especially in real-time applications. While state-of-the-art models often attain impressive accuracy, their deployment on resource-constrained devices is limited by substantial latency and memory overhead. This trade-off is particularly critical in embedded and edge scenarios, where computational resources are inherently limited. In this study, we introduce a computation-efficient model specifically designed for deployment on such low-computation devices without resorting to architecture search techniques like pruning or additional training procedures such as knowledge distillation. Instead, the proposed model exhibits strong performance under low-precision inference settings, made possible through model compression via quantization (Sec.~\ref{deployment}). Our architecture integrates a carefully optimized combination of dilated convolutions and depthwise separable convolutions within the encoder, balancing representational capacity and computational efficiency. This design not only preserves high segmentation accuracy but also achieves low inference latency, rendering it well-suited for real-time deployment in practical environments.

\begin{figure*}[!t]
\centering
\includegraphics[width=1\linewidth]{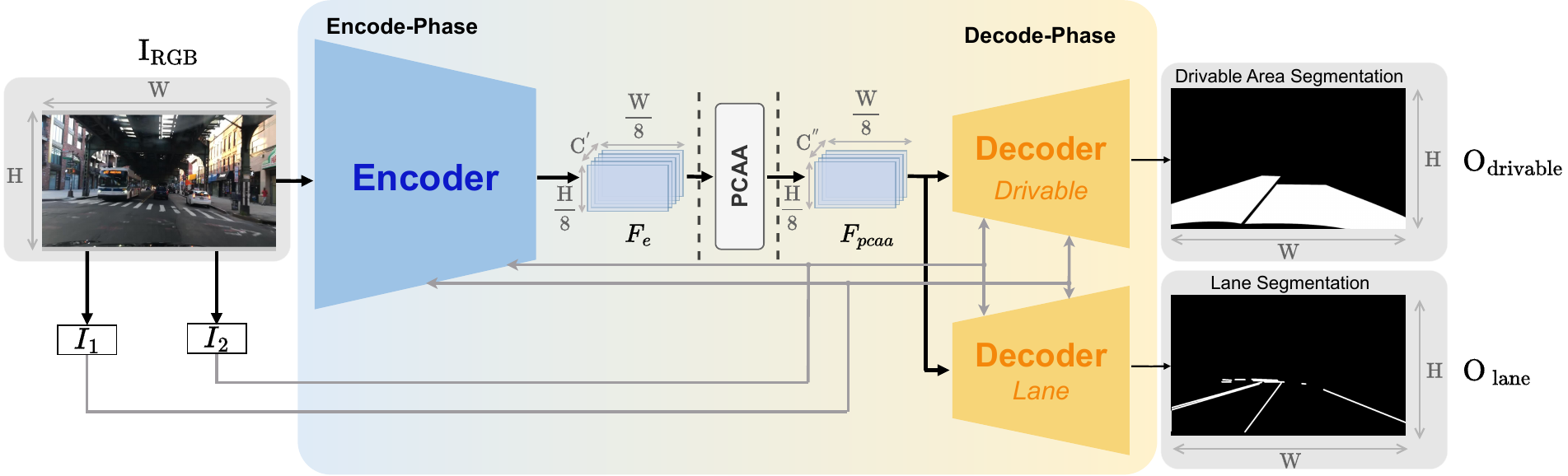}
\caption{The TwinLiteNet$^+$ architecture comprises two phases. During the Encode phase, the input image passes through an Encoder block followed by a Partial Class Activation Attention mechanism. In the Decode phase, the output from the Encoder is channeled through two identical yet independent Decoder blocks, transforming the feature maps into two separate segmentation maps.}    
\label{arch}
\end{figure*}

\algrenewcommand\algorithmicrequire{\textbf{Input:}}
\algrenewcommand\algorithmicensure{\textbf{Output:}}

\section{\textbf{Proposed method}} \label{Propose}

This paper introduces TwinLiteNet$^+$, a model specifically engineered for segmenting drivable areas and lanes. Inspired by the preceding TwinLiteNet \cite{twin}, our study aims to enhance the model by refining both the encoder and decoder components. In particular, the proposed decoder is designed to effectively leverage information compared to its predecessor while still maintaining a low computational resource utilization. Moreover, the encoder integrates depthwise dilated separable convolutions, thereby optimizing the inference time of the model which is suitable for real-time implementation. TwinLiteNet$^+$ operates in two primary phases: the Encode-phase and the Decode-phase, incorporating a Partial Class Activation Attention (PCAA) module \cite{pcaa} which enhances segmentation precision and efficiency by focusing on key areas like drivable zones and lanes. While several attention mechanisms such as DANet \cite{cfam}, SE \cite{senet}, and CBAM \cite{cbam} have been proposed to enhance feature representations by modeling spatial and/or channel-wise dependencies, they typically rely on global feature statistics or self-attention across the entire feature map without explicitly incorporating semantic class information. In contrast, PCAA generates attention maps guided by partial class activation maps, which embed category-level semantics. This enables PCAA to focus selectively on regions relevant to specific classes, offering both strong discriminative capability and high computational efficiency.

In the Encoding stage, the model processes the input image $\textbf{I}_\text{RGB} \in \mathbb{R}^{3 \times H \times W}$ using a shared-weight Encoder block, designed to extract pertinent image features. This block leverages an extended receptive field provided by dilated convolutions, combined with the low computational demand yet high efficiency of Depthwise Convolution, thereby enhancing the model's performance while maintaining low latency. The Encoder's output $\textbf{F}_\text{e} \in \mathbb{R}^{C^{'} \times \frac{H}{8} \times \frac{W}{8}}$ is processed by the PCAA mechanism, emphasizing key features, particularly of drivable areas and lanes. PCAA operates by collecting local class representations based on partial Class Activation Maps (CAMs) and calculating pixel-to-class similarity maps within patches, a significant approach for advancing the precision and effectiveness of semantic segmentation models. Following this, a feature map $\textbf{F}_\text{pcaa} \in \mathbb{R}^{C^{''} \times \frac{H}{8} \times \frac{W}{8}}$ is channeled into two separate Decoder blocks, each tasked with specific prediction objectives. The output of these blocks yields segmentation maps for Drivable Area Segmentation and Lane Segmentation, represented as $\textbf{O}_\text{drivable}$, $\textbf{O}_\text{lane} \in \mathbb{R}^{2 \times H \times W}$.The simple feedforward process is outlined in Algorithm \ref{feedforward}. The model’s training integrates both Focal \cite{focal} and Tversky loss \cite{tversky} functions. TwinLiteNet$^+$ is available in four configurations: Nano, Small, Medium to Large, offering adaptability based on the computational power of the underlying hardware. The comprehensive architecture of TwinLiteNet$^+$ is depicted in Figure \ref{arch}.

\begin{figure*}[t]
  \centering
    \begin{subfigure}{0.30\columnwidth}
        \includegraphics[width=\textwidth]{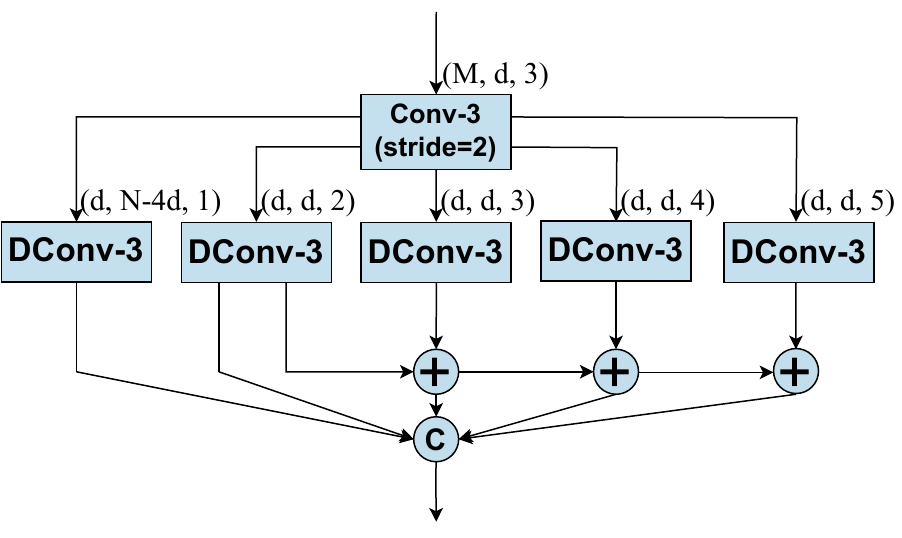}
        \caption{Stride ESP}
        \label{esp1}
    \end{subfigure} 
    \begin{subfigure}{0.34\columnwidth}
        \includegraphics[width=\textwidth]{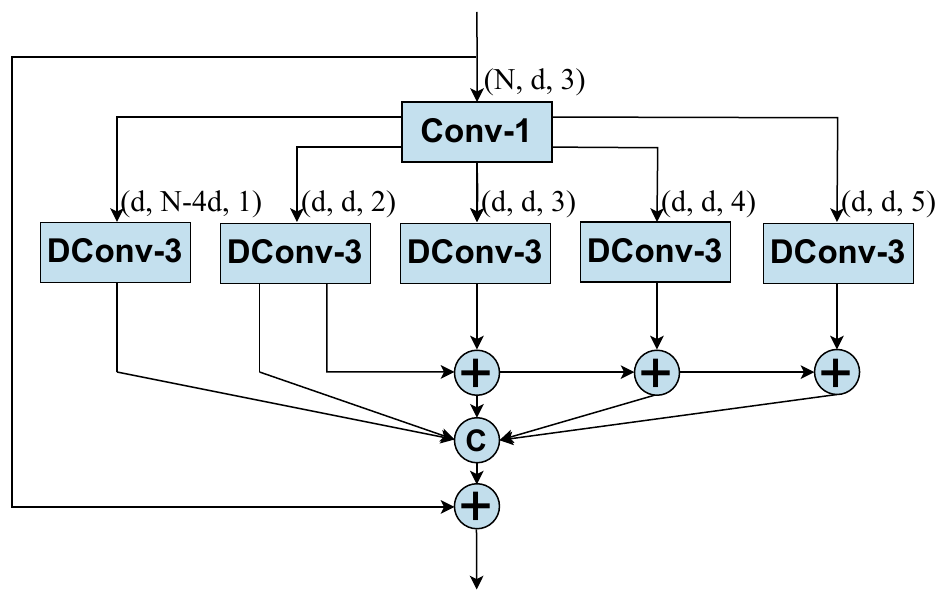}
        \caption{ESP}
        \label{esp2}
    \end{subfigure} 
    \begin{subfigure}{0.34\columnwidth}
        \includegraphics[width=\textwidth]{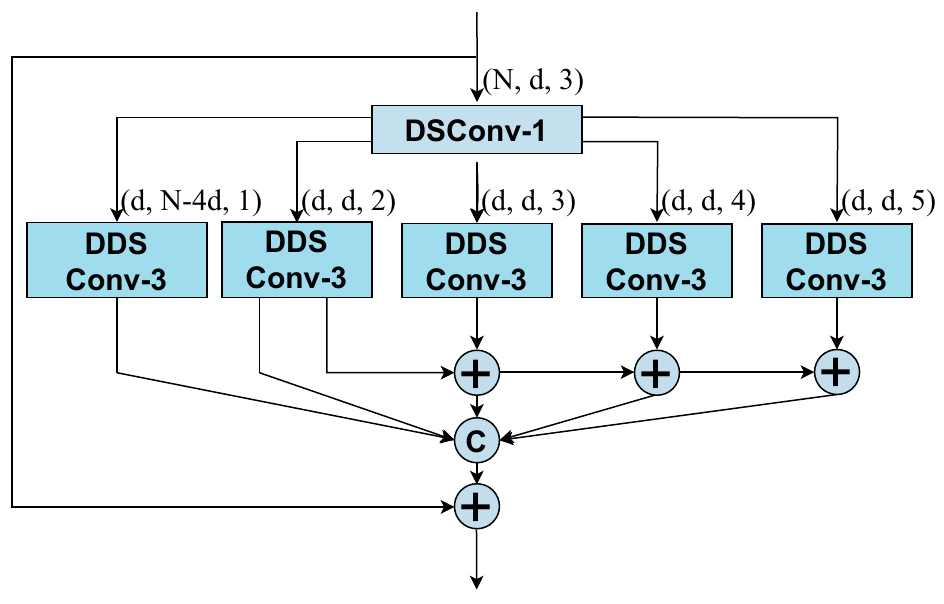}
        \caption{DESP} 
        \label{esp3}
    \end{subfigure} 
    \caption{This figure presents variants of ESP blocks within the Encoder. The convolutional layers are designated as Conv-n (standard $n \times n$ convolution), DConv-n ($n \times n$ dilated convolution) in Stride ESP (Fig \ref{esp1}) and ESP (Fig \ref{esp2}), and DSConv-n ($n \times n$ depthwise separable convolution), DDSConv-n ($n \times n$ depthwise dilated separable convolution) in Stride DESP (Fig \ref{esp3}), and are described in terms of $(input$ $channels,$ $output$ $channels,$ $dilation$ $level)$, where $dilation$ $rate$ = $2^{\text{\textit{dilation level}}-1}$. These blocks consolidate feature maps employing a Hierarchical Feature Fusion (HFF)\cite{espnet}, which is notable for its computational efficiency and its ability to eliminate grid-like artifacts resulting from dilated convolutions. The Stride ESP (a) not only accomplishes downsampling but also converts the depth of the feature map from $M$ to $N$, whereas ESP (Fig \ref{esp2}) and DESP (Fig \ref{esp3}) maintain the original dimensions of the feature map. The main difference between the ESP and DESP modules lies in the use of depthwise dilated separable convolution (DDSConv-n) layers in DESP instead of dilated convolution (DConv-n). In this design, each depthwise dilated separable convolution layer applies depthwise dilated convolutions with different $dilation$ $rate$ to extract spatial features, followed by a pointwise convolution to fuse information across channels.}
    \label{esp}
\end{figure*}

\begin{algorithm}[!b]
    \caption{Feedforward process of TwinLiteNet$^+$}\label{feedforward}

    \begin{algorithmic}[1]
    \Require The input image \textbf{I}$_\text{RGB}$ and two downsampled image \textbf{I}$_1$ and \textbf{I}$_2$ \Comment{\textit{\textbf{I}$_1$ and \textbf{I}$_2$ are downsampling images derived from equations \ref{1} and \ref{2}, respectively.}}
    \Ensure Drivable Area Segmentation Map $\textbf{O}_\text{drivable}$ and Lane Segmentation Map $\textbf{O}_\text{lane}$
        \State \textbf{F}$_\textbf{e}$ $\gets$ Encoder(\textbf{I}$_\text{RGB}$, \textbf{I}$_{1}$, I$_{2}$) 
        \State \textbf{F}$_\text{pcaa}$ $\gets$ PCAA(F$_\text{e}$)
        \State \textbf{O}$_\text{drivable}$ $\gets$ Decoder$_\text{drivable}$(\textbf{I}$_\text{pcaa}$, \textbf{I}$_{1}$, \textbf{I}$_{2}$)
        \State \textbf{O}$_\text{lane}$ $\gets$ Decoder$_\text{lane}$(\textbf{I}$_\text{pcaa}$, \textbf{I}$_{1}$, \textbf{I}$_{2}$)
    \State \Return \textbf{O}$_\text{lane}$, \textbf{O}$_\text{drivable}$
    \end{algorithmic}
\end{algorithm}

\subsection{\textbf{Encoder}}

In the development of the TwinLiteNet$^+$ model, we innovate an efficient and computationally cost-effective encoder, consisting of a series of convolutional layers designed for extracting features from input images. This encoder draws inspiration from the ESPNet encoder \cite{espnet}, a modern and efficient network for segmentation tasks. The cornerstone of the ESPNet architecture is the ESP module, depicted in Figure \ref{esp2}. The ESP unit initially projects the high-dimensional input feature map into a lower-dimensional space using point-wise convolutions (or $1 \times 1$ convolutions), subsequently learning parallel representations via dilated convolutions with varying dilation rates. This differs from the use of standard convolutional kernels. The ESP module undergoes several stages, including Reduce, Split, Transform, and Merge, with its detailed implementation illustrated in Figure \ref{esp}. For computations with the ESP module, the feature map first undergoes down-sampling in the Stride ESP block, where point-wise convolutions are replaced with n$\times$n stride convolutions within the ESP module to enable nonlinear down-sampling operations, shown in Figure \ref{esp1}. The spatial dimensions of the feature map are altered through down-sampling operations, changing from $\textbf{F}_\text{in} \in \mathbb{R}^{M \times H' \times W'} \to \textbf{F}_{0} \in \mathbb{R}^{N \times \frac{H'}{2} \times \frac{W'}{2}}$. The ESP module in ESPNet iterates output of Stride ESP to increase the network's depth, transforming $\textbf{F}_{i-1} \in \mathbb{R}^{N \times \frac{H'}{2} \times \frac{W'}{2}} \to \textbf{F}_{i} \in \mathbb{R}^{N \times \frac{H'}{2} \times \frac{W'}{2}}$. During this phase, the feature map maintains its size and depth, thus the input and output feature maps of the ESP module are combined using element-wise summation to enhance information flow. The ESP and Stride ESP modules are followed by Batch Normalization \cite{bn} and PReLU \cite{prelu} nonlinearity. The final output, $\textbf{F}_\text{out} \in \mathbb{R}^{2N \times \frac{H'}{2} \times \frac{W'}{2}}$, is obtained through a concatenation operation between F$_0$ and F$_\text{N}$, effectively expanding the feature map's dimensions. This process is detailed in Equation \ref{phase1}.

\begin{figure}[t]
\centering
\includegraphics[width=1.\linewidth]{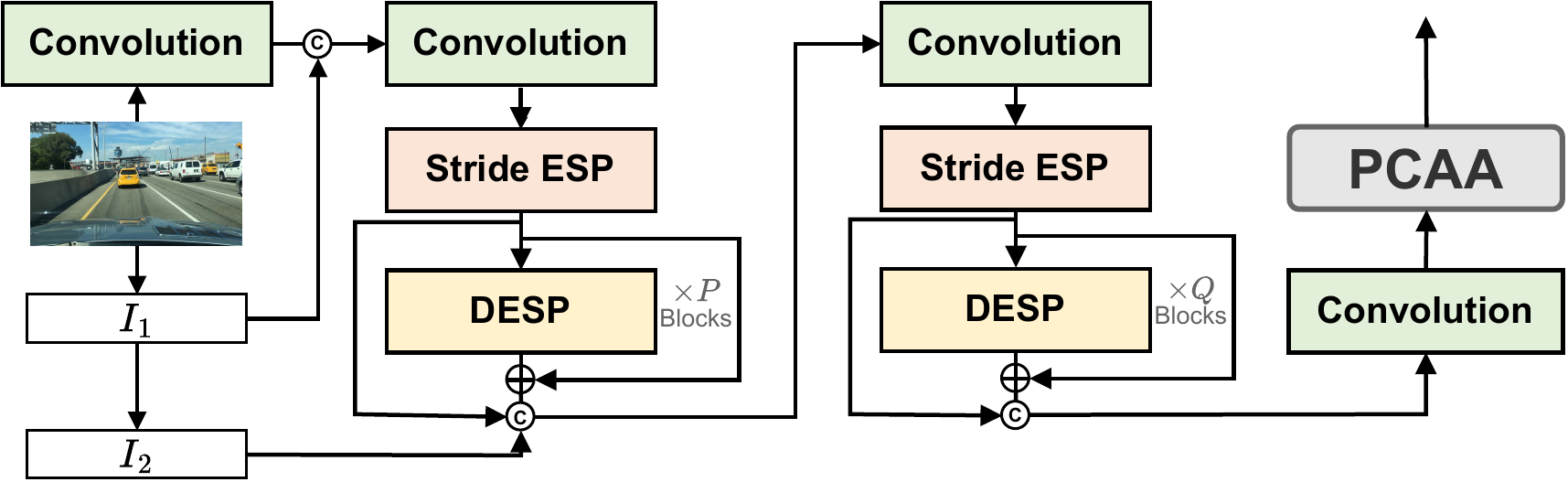}
\caption{Comprehensive schematic of the Encoder in TwinLiteNet$^+$.}    
\label{encoder}
\end{figure}

\begin{equation}
\label{phase1}
\begin{split}
\begin{cases}
		\textbf{F}_{0} = \text{StrideESP}(\textbf{F}_\text{in})\\
        \textbf{F}_{i} = \text{ESP}(\textbf{F}_{i-1}) + \textbf{F}_{i-1}, & \text{$i \in \left[ 1,N \right]$}\\
        \textbf{F}_{out} = Concat(\textbf{F}_\text{0}, \textbf{F}_\text{N})
\end{cases}
\end{split}
\end{equation}

As previously addressed in Section \ref{2c}, while standard dilated convolutions achieve enhanced accuracy, they also bear significantly higher computational costs compared to their depthwise separable counterparts. So, in the TwinLiteNet$^+$ model, we propose using Depthwise ESP (DESP) as an alternative to the ESP module, while maintaining the Stride ESP module. The DESP is adeptly engineered to substitute standard dilated convolutional layers with depthwise separable dilated convolutional layers. This proposed approach preserves the efficacy of the Stride ESP module while markedly reducing computational expenses by substituting the ESP module with DESP, as depicted in Figure \ref{esp3}. For instance, in a feature map sized $64 \times 180 \times 320$, the DESP algorithm necessitates learning merely 2,332 parameters, incurring a computational cost of 0.14 GFLOPs. Conversely, the ESP algorithm requires learning a considerably greater number of parameters, amounting to 7,872, with a related computational cost of 0.46 GFLOPs. In our proposed model, the ESP module is iterated many times. Using DESP to replace ESP reduces the model's parameters and computational costs. The computational procedures of the Stride ESP and DESP blocks are executed as outlined in Eq. \ref{phase2}.

\begin{equation}
\label{phase2}
\begin{split}
\begin{cases}
		\textbf{F}_{0} = \text{StrideESP}(\textbf{F}_\text{in})\\
        \textbf{F}_{i} = \text{DESP}(\textbf{F}_{i-1}) + \textbf{F}_{i-1}, & \text{$i \in \left[ 1,N \right]$}\\
        \textbf{F}_{out} = Concat(\textbf{F}_\text{0}, \textbf{F}_\text{N})
\end{cases}
\end{split}
\end{equation}
Moreover, our encoder block introduces efficient long-range shortcut connections between the input image and the current downsampling unit, thereby enhancing the encoding of spatial relationships and facilitating more effective representation learning. These connections initially downsample the image to align with the feature map dimensions using Average Pooling. Our encoder block consists of two downsampled images, $\textbf{I}_{1} \in \mathbb{R}^{3 \times \frac{H}{2} \times \frac{W}{2}}$ and $\textbf{I}_{2} \in \mathbb{R}^{3 \times \frac{H}{4} \times \frac{W}{4}}$ (Eqs. \ref{1} – \ref{2}). Figure \ref{encoder} depicts the proposed Encoder Block in detail. Our Stride ESP and DESP modules are cohesively integrated via convolutional layers, succeeded by batch normalization \cite{bn} and PReLU non-linearity \cite{prelu}. In the Encoder model, two computations are performed by combining Stride ESP and DESP blocks as calculated in Eq. \ref{phase2}. In the first calculation, after the Stride ESP block computes the solution, the DESP block will execute \textbf{P} times instead of $N$. After executing the DESP blocks, the final output is concatenated with the output of Stride ESP and $\textbf{I}_{2}$ before being sent to the following calculation phase. Meanwhile, the DESP blocks in the second calculation are executed similarly to the first calculation, with the DESP blocks being executed \textbf{Q} times, and the final output when executing the DESP blocks is concatenated with the Stride ESP in the same calculation. The hyperparameters \textbf{P} and \textbf{Q} are determined based on the selected model config, and further specifics are delineated in Section \ref{configsec}.

\begin{equation}
\label{1}
    \textbf{I}_1 = AvgPool(\textbf{I}_\text{RGB})
\end{equation}
\begin{equation}
\label{2}
    \textbf{I}_2 = AvgPool(AvgPool(\textbf{I}_\text{RGB}))
\end{equation}

\begin{figure}[t]
\centering
\includegraphics[width=0.9\linewidth]{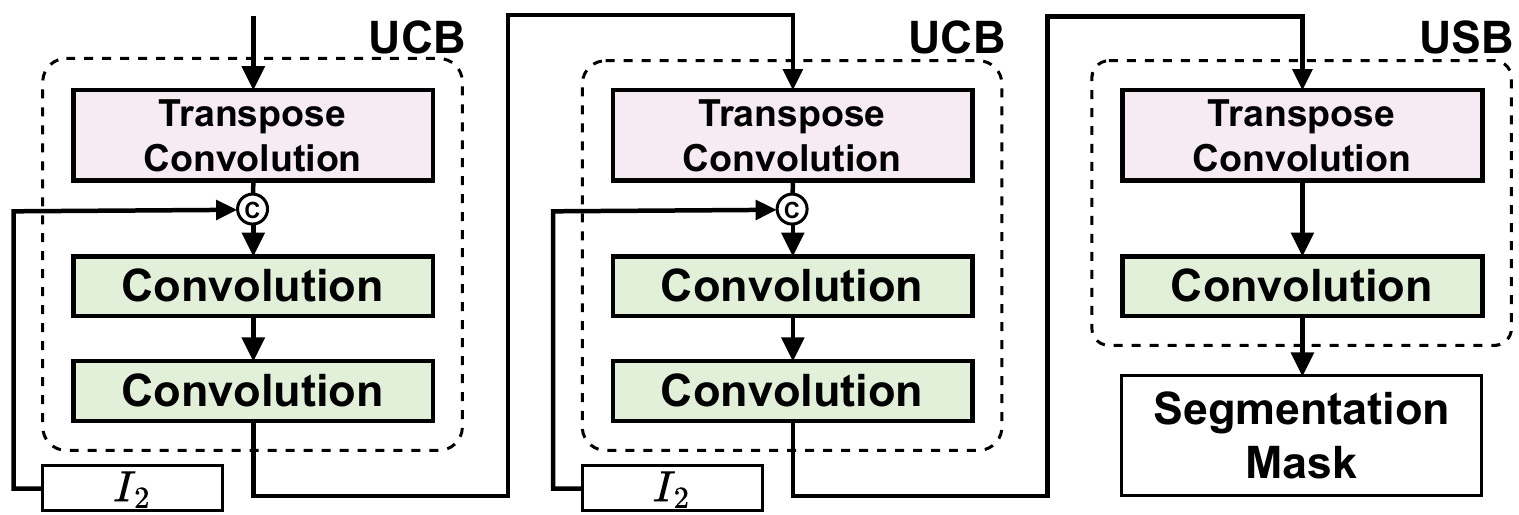}
\caption{Decoder Block Design in TwinLiteNet$^+$. This illustrates the implementation of Upper Convolution Block (UCB) and Upper Simple Block (USB) within the decoder, specifically tailored for upsampling to generate segment maps for diverse tasks.}    
\label{decoder}
\end{figure}

\subsection{\textbf{Decoder}}

In our TwinLiteNet$^+$ architecture, we implement an innovative approach by deploying multiple decoding modules, each tailored for a distinct segmentation task. This design deviates from conventional methods that typically depend on a singular output for all object categories, with the tensor output encompassing $C+1$ channels, corresponding to $C$ classes and one additional channel for the background. Specifically, the post-processing feature map obtained from PCAA denoted as $\textbf{F}_{pcaa} \in \mathbb{R}^{C'' \times \frac{H}{8} \times \frac{W}{8}}$, our architecture employs two separate yet structurally identical decoding blocks. These blocks independently handle the segmentation of distinct regions and drivable lanes. Our decoder effectively reduces the depth of the feature map and employs upsampling techniques. The upsampling process within the decoder is facilitated by a combination of Transpose Convolution and Convolution operations. We introduce two novel blocks, namely the Upper Convolution Block (UCB) and Upper Simple Block (USB), both elegantly designed for upsampling purposes. The UCB comprises a Transpose Convolution followed by two subsequent Convolution operations, wherein the post-Transpose Convolution features are concatenated with the input image at a lower resolution, either I$_1$ or I$_2$. Conversely, the USB incorporates a single Transpose Convolution and one Convolution, without merging input image information, and its output constitutes the segment map as predicted by the model. We apply batch normalization and PReLU following each Transpose Convolution or Convolution layer. Integrating a Convolution layer immediately subsequent to a Transpose Convolution significantly enhances the feature learning capabilities of the Decoder, thereby facilitating the restoration of resolution lost during the downsampling process in the encoder. The detailed architecture of our model's Decoder Block is depicted in Figure \ref{decoder}. We have strategically designed our decoder to be simplistic, thereby ensuring minimal computational overhead for the model. The sequential application of the blocks $UCB \to UCB \to USB$ leads to a systematic change in the feature map dimensions from $\frac{H}{8} \times \frac{W}{8} \to \frac{H}{4} \times \frac{W}{4} \to \frac{H}{2} \times \frac{W}{2} \to H \times W$ for each targeted task. This modular, branched approach allows for the independent tuning of each task, facilitating the creation of optimal conditions for refining specific segmentation outcomes for various regions and drivable lanes. As a result, our model effectively differentiates and enhances features relevant to each segmentation task, culminating in an output resolution of $\textbf{O} \in \mathbb{R}^{2 \times H \times W}$.

\subsection{\textbf{Model configurations}} \label{configsec}

To ensure versatility and broad applicability across diverse hardware devices, we design our TwinLiteNet$^+$ model with four configurations: \textit{Nano}, \textit{Small}, \textit{Medium}, and \textit{Large}. These variants, denoted as TwinLiteNet$^+_{\text{Nano}}$, TwinLiteNet$^+_{\text{Small}}$, TwinLiteNet$^+_{\text{Medium}}$, and TwinLiteNet$^+_{\text{Large}}$, differ primarily in the number of convolutional filters and the structural hyperparameters \textbf{P} and \textbf{Q}, while preserving the overall encoder-decoder architecture. This configuration hierarchy allows for a flexible trade-off between performance and computational cost. Specifically, the Nano version contains only 0.03 million parameters and requires 0.57 GFLOPs, making it highly suitable for ultra-low-power edge devices. In contrast, the Large version reaches 1.94 million parameters and 17.58 GFLOPs, offering improved accuracy at the cost of higher resource consumption. All configurations share the same architectural stages but scale their complexity by adjusting the output channel widths at each stage. The encoder stages consist of stacked convolutional blocks with downsampling via Stride ESP and a repeated application of DESP determined by the hyperparameters \textbf{P} and \textbf{Q}, which control the depth and receptive field of the network. Table~\ref{configtab} provides a comprehensive overview of the architectural parameters and resource consumption across different model configurations.

This modular scaling strategy enables our model to adapt to varying deployment constraints without requiring changes in architecture design or retraining. In particular, the lightweight TwinLiteNet$^+_{\text{Nano}}$ proves highly effective in latency-sensitive applications, while the Medium and Large configurations deliver stronger performance in scenarios with more relaxed computational budgets.

\begin{table}[]
\centering
\setlength{\tabcolsep}{20pt}
\caption{Model architecture settings in different TwinLiteNet$^+$ model's configs: Nano, Small, Medium and Large.}
\label{configtab}
% \resizebox{\textwidth}{!}{
\begin{tabular}{lccccc}
\hline & \\[\dimexpr-\normalbaselineskip+3pt]
\multirow{2}{*}{\textbf{Layer}} & \multirow{2}{*}{\textbf{Ouput Size}}              & \multicolumn{4}{c}{\textbf{Output channels for TwinLiteNet$^+$}}    \\ \cline{3-6} 
                                &                                                   & \textbf{Nano} & \textbf{Small} & \textbf{Medium} & \textbf{Large} \\ & \\[\dimexpr-\normalbaselineskip+3pt] \hline & \\[\dimexpr-\normalbaselineskip+3pt]
Conv                             & \multirow{2}{*}{$\frac{H}{2} \times \frac{W}{2}$} & 4             & 8              & 16              & 32             \\
Conv                             &                                                   & 8             & 16             & 32              & 64             \\ & \\[\dimexpr-\normalbaselineskip+3pt] \hline & \\[\dimexpr-\normalbaselineskip+3pt]
Stride ESP                             & \multirow{3}{*}{$\frac{H}{4} \times \frac{W}{4}$} & 16            & 32             & 64              & 128            \\
P $\times$ DESP                 &                                                   & 16            & 32             & 64              & 128            \\
Conv                             &                                                   & 32            & 64             & 128             & 256            \\ & \\[\dimexpr-\normalbaselineskip+3pt] \hline & \\[\dimexpr-\normalbaselineskip+3pt]
Stride ESP                             & \multirow{5}{*}{$\frac{H}{8} \times \frac{W}{8}$} & 32            & 64             & 128             & 256            \\
Q $\times$DESP                  &                                                   & 32            & 64             & 128             & 256            \\
Conv                             &                                                   & 16            & 32             & 64              & 128            \\
PCAA                            &                                                   & 16            & 32             & 64              & 128            \\
Conv                             &                                                   & 8             & 16             & 32              & 64             \\ & \\[\dimexpr-\normalbaselineskip+3pt] \hline & \\[\dimexpr-\normalbaselineskip+3pt]
\begin{tabular}[c]{@{}l@{}}UCB$_\text{drivable}$\\ UCB$_\text{lane}$\end{tabular}                               & $\frac{H}{4} \times \frac{W}{4}$                  & 4             & 8              & 16              & 32             \\ & \\[\dimexpr-\normalbaselineskip+3pt] \hline & \\[\dimexpr-\normalbaselineskip+3pt]
\begin{tabular}[c]{@{}l@{}}UCB$_\text{drivable}$\\ UCB$_\text{lane}$\end{tabular}                               & $\frac{H}{2} \times \frac{W}{2}$                  & 4             & 8              & 8               & 8              \\ & \\[\dimexpr-\normalbaselineskip+3pt] \hline & \\[\dimexpr-\normalbaselineskip+3pt]
\begin{tabular}[c]{@{}l@{}}USB$_\text{drivable}$\\ USB$_\text{lane}$\end{tabular}                              & $H \times W$                                      & 2             & 2              & 2               & 2              \\ & \\[\dimexpr-\normalbaselineskip+3pt] \hline \hline & \\[\dimexpr-\normalbaselineskip+3pt]

\multicolumn{2}{l}{\textbf{P}, \textbf{Q}}                                                            & 1, 1          & 2, 3           & 3, 5            & 5, 7           \\ \hline
\multicolumn{2}{l}{\#Param.}                                                        & 0.03M         & 0.12M          & 0.48M           & 1.94M          \\ \hline
\multicolumn{2}{l}{FLOPs}                                                           & 0.57G         & 1.40G          & 4.63G           & 17.58G          \\ \hline
\end{tabular}
% }
\end{table}

\subsection{\textbf{Loss Function}}

To effectively address the specific challenges posed by pixel-wise classification in autonomous driving scenarios-particularly class imbalance and sensitivity to fine structural details-we employ a combination of Focal Loss \cite{focal} and Tversky Loss \cite{tversky}. Each loss is applied independently to both segmentation heads to target complementary aspects of the task.

\textbf{Focal Loss} \cite{focal}: Designed to down-weight well-classified examples and focus learning on harder, misclassified pixels, Focal Loss is particularly beneficial in scenarios with significant class imbalance-such as drivable area and lane segmentation, where background pixels dominate. For example, in lane segmentation, narrow foreground structures (i.e., lanes) are often underrepresented, making them prone to being overlooked by standard losses. Focal Loss addresses this by introducing a modulating factor $(1 - \hat{p}_i(c))^\gamma$ to the standard cross-entropy:

\begin{equation}
\label{focal}
\mathcal{L}_{focal} = -\frac{1}{N} \sum_{c=0}^{C-1} \sum_{i=1}^{N} p_i(c)(1 - \hat{p}_i(c))^\gamma \log(\hat{p}_i(c))
\end{equation}
where, $N$ is the number of pixels, $C = 2$ is the number of classes (foreground vs. background), $p_i(c)$ is the one-hot ground truth label for pixel $i$ and class $c$, $\hat{p}_i(c)$ is the predicted probability, and $\gamma$ is a focusing parameter (commonly set to 2) that reduces the impact of easy examples.

\textbf{Tversky Loss} \cite{tversky}: An extension of Dice Loss \cite{dice}, Tversky Loss introduces tunable parameters $\alpha$ and $\beta$ to balance penalties on false positives and false negatives. This is particularly important in lane segmentation, where missing thin, elongated structures is more critical than predicting false positives. The formulation is:

\begin{equation}
\label{tversky}
\mathcal{L}_{tversky} = \sum_{c=0}^{C-1} \left(1 - \frac{TP(c)}{TP(c) + \alpha FN(c) + \beta FP(c)}\right)
\end{equation}
where $TP$, $FN$, and $FP$ denote true positives, false negatives, and false positives, respectively. Choosing $\alpha > \beta$ encourages the model to prioritize recall-an essential property for safety-critical applications like lane segmentation.

\textbf{Total Loss:} The total loss combines the drivable area and lane-specific losses:

\begin{equation}
\label{losstotal}
\mathcal{L} = \mathcal{L}_{\text{drivable}} + \mathcal{L}_{\text{lane}}
\end{equation}

Each sub-loss is a weighted combination of the Focal and Tversky losses applied to the corresponding output. Focal Loss mitigates class imbalance by encouraging the model to focus on harder pixels, while Tversky Loss enhances robustness in detecting thin structures by emphasizing recall. Together, they form a complementary loss framework that improves both the robustness and accuracy of segmentation in complex real-world driving environments.

\begin{algorithm}[t]

\caption{Training and evaluation process for the proposed TwinLiteNet$^+$ model}
\label{train}
\KwIn{Target end-to-end network $F$ with parameters $\Theta$; Training dataset $\tau_{train}$; Validation dataset $\tau_{val}$}
% \KwOut{Well-trained network $F(x; \Theta)$}
Initialize the parameters $\Theta$ \;
\For{$epoch \leftarrow 1$ \KwTo $epochs$}{
  \For{each batch $(x_{train}, y_{train})$ in $\tau_{train}$}{
    $y_{drivable}, y_{lane} \gets y_{train}$ \;
    $\hat{y}_{drivable}, \hat{y}_{lane} \gets F(x_{train})$ \;
    $\mathcal{L}_{drivable} \gets \mathcal{L}^{focal}_{drivable} + \mathcal{L}^{tversky}_{drivable}$ \;
    $\mathcal{L}_{lane} \gets \mathcal{L}^{focal}_{lane} + \mathcal{L}^{tversky}_{lane}$ \;
    $\mathcal{L} \gets \mathcal{L}_{drivable} + \mathcal{L}_{lane}$ \;
    $\Theta \gets \arg\min_{\Theta}\mathcal{L}$ \;
    $\Theta \gets \text{Update}_{EMA}(\Theta)$ \;
  }
}
$\hat{y}_{val} \gets F(x_{val}; \Theta)$ \;
$mIoU_{drivable}, Acc_{lane}, IoU_{lane} \gets Evaluate(\hat{y}_{val}, y_{val})$ \;
\Return $mIoU_{drivable}, Acc_{lane}, IoU_{lane}$
\end{algorithm}

\section{\textbf{Experimental}} \label{result}
\subsection{\textbf{Experiment Setup}}
\subsubsection{\textbf{Dataset}}
The BDD100K \footnote{\url{https://bdd-data.berkeley.edu}} dataset was used for training and validating TwinLiteNet$^+$. With 100,000 frames and annotations for 10 tasks, it is a large dataset for autonomous driving. Due to its diversity in geography, environment, and weather conditions, the algorithm trained on the BDD100K dataset is robust enough to generalize to new settings. The BDD100K dataset is divided into three parts: a training set with 70,000 images, a validation set with 10,000 images, and a test set with 20,000 images. Since no labels are available for the 20,000 images in the test set, we choose to evaluate on a separate validation set of 10,000 images. We use data that is prepared similarly to several previous studies \cite{enetsad,twin} to ensure fairness in comparison.

\subsubsection{\textbf{Evaluation Metrics}}
For the segmentation tasks, similar to the evaluation approach in \cite{twin,BILane,interactive}, we assess the drivable area segmentation task using the mIoU (mean Intersection of Union) metric. In the context of lane segmentation, we measure the performance using both accuracy and IoU (Intersection over Union) metrics. However, owing to pixel imbalances between the background and foreground in lane segmentation, we opt for a more meaningful balanced accuracy metric in our evaluation. Traditional accuracy metrics may produce biased results by favoring classes with a larger number of samples. In contrast, balanced accuracy provides a fairer assessment by considering the accuracy for each class. We use the accuracy metrics provided in the study \cite{yolom}. To evaluate computational efficiency, we assess the model using two standard metrics: parameters and FLOPs. In line with established conventions in the field \cite{resnet,espnet}, we use FLOPs as the total number of multiply-add operations needed for inference. 

\subsubsection{\textbf{Implementation details}}
To improve performance, we apply several data augmentation techniques. To address photometric distortions, we modify the hue, saturation, and value parameters of the image. In addition, we also incorporate basic enhancement techniques to handle geometric distortions such as random translation, cropping, and horizontal flipping.
We train our model using the AdamW \cite{adamw} optimizer with a learning rate of $5 \times 10^{-4}$, momentum of 0.9, and weight decay of $5 \times 10^{-4}$. In our TwinLiteNet$^+$, we use Exponential Moving Average (EMA) model \cite{ema} purely as the final inference model. Additionally, we resize the original image dimensions from 1280$\times$720 to 640$\times$384. For the loss function coefficients, we set $\alpha$ = 0.7 and $\beta$ = 0.3 for $\mathcal{L}_{drivable}^{tversky}$, $\alpha$ = 0.9 and $\beta$ = 0.1 for $\mathcal{L}_{lane}^{tversky}$, and $\alpha_t$ = 0.25 and $\gamma$ = 2 in $\mathcal{L}_{drivable, lane}^{focal}$. All of the specified configurations are developed based on empirical evidence. Finally, we conduct training with a batch size of 16 on an RTX A5000 for 100 epochs. The model training and evaluation process is described in Algorithm \ref{train}. During the model training process, we simultaneously train both tasks and employ EMA technique for weight updates following each backpropagation step. After each epoch, we evaluate the model on the validation set and focus on three key metrics: mean Intersection over Union for the Drivable Area Segmentation task (mIoU), Lane Accuracy (Acc), and Intersection over Union (IoU) for the Lane Segmentation task. All experiments used the PyTorch framework on an NVIDIA GeForce RTX A5000 GPU with 32GB of RAM and an Intel(R) Core(TM) i9-10900X processor.

\subsection{\textbf{Main results}}

\subsubsection{\textbf{Cost Computation Performance}}

\begin{table*}[h]
\centering
\caption{Comparison between TwinLiteNet$^+$ Model's configs in FPS (in 5 different batch sizes), Parameters, FLOPs and Model size. (Parameters and FLOPs represent the number of parameters and the number of floating point operations required in each model, FPS shows the model's latency when inference.}
\label{cost}
\begin{tabular}{lcccccccc}
\toprule
\multirow{2}{*}{\textbf{Config}} & \multicolumn{5}{c}{\textbf{FPS (with Batch Size)} $\uparrow$} & \multirow{2}{*}{\textbf{\#Param.} $\downarrow$} & \multirow{2}{*}{\begin{tabular}[c]{@{}c@{}}\textbf{FLOPs} $\downarrow$\\ \textbf{(batch=1)}\end{tabular}} & \multirow{2}{*}{\textbf{Model Size} $\downarrow$} \\ \cmidrule{2-6}
                       & \textbf{1}        & \textbf{2}        & \textbf{4}        & \textbf{8}        & \textbf{16}       &                                       &                                                                                         &                                          \\ \midrule
Nano   & 237      & 470      & 921      & 1004     & 1163     & 0.03M                                 & 0.57G                                                                                   & 0.06MB                                   \\
Small  & 178      & 340      & 697      & 732      & 779      & 0.12M                                 & 1.40G                                                                                   & 0.23MB                                   \\
Medium & 138      & 269      & 442      & 456      & 484      & 0.48M                                 & 4.63G                                                                                   & 0.92MB                                   \\
Large  & 109      & 186      & 218      & 220      & 237      & 1.94M                                 & 17.58G                                                                                  & 3.72MB                                   \\ \bottomrule
\end{tabular}
\end{table*}

Table \ref{cost} presents the computational cost outcomes of various configs of the TwinLiteNet$^+$ model. The parameters include the number of parameters ($\#$Param.), and FLOPs indicating the computation required per inference (with a batch size of 1), along with the model's size. Frame rate (FPS) is measured across different batch sizes (1, 2, 4, 8, 16) to assess latency during inference. The results were obtained while doing inference with PyTorch FP32 (without TorchScript, or TensorRT). Results show that TwinLiteNet$^+_{\text{Nano}}$ is the lightest config with only 0.03M parameters and 0.57G FLOPs, achieving an impressive inference speed of up to 1163 FPS with a batch size of 16. Inference with larger batch sizes opens up the potential for simultaneous inference across multiple cameras, a critical capability for autonomous vehicle applications. However, transitioning from the Nano to the Large config, the parameters increase up to 1.94M and FLOPs to 17.58G, indicating a more powerful computational capability but also resulting in an increase in latency, with inference speed dropping to 237 FPS. The trade-off between accuracy and latency is evident: larger models offer higher computational potential but require more processing time. This highlights the importance of carefully balancing the need for accuracy and real-time requirements when selecting a model for specific applications.

\begin{table*}[!t]
\caption{Performance benchmarking in mIoU (\%) for Drivable Area Segmentation, IoU (\%) and Acc (\%) for Lane Segmentation of models with same tasks. $^{\ast}$ denotes results directly cited from \cite{sparse}. The best and second best results are marked in \textbf{bold} and \underline{underline} respectively.}
  \centering
\label{tab: dall}
\resizebox{\columnwidth}{!}{
\begin{tabular}{lcccccc}
\toprule
\multirow{2}{*}{\textbf{Model}}                                                                     & \textbf{Drivable Area}        &  & \multicolumn{2}{c}{\textbf{Lane}}             & \multirow{2}{*}{\textbf{FLOPS} $\downarrow$} & \multirow{2}{*}{\textbf{\#Param.} $\downarrow$} \\ \cline{2-2} \cline{4-5}
                                                                                           & \textbf{mIoU (\%)} $\uparrow$ &  & \textbf{Acc (\%)} $\uparrow$ & \textbf{IoU (\%)} $\uparrow$ &                                     &                                      \\ \midrule
DeepLabV3+$^{\ast}$ \cite{deeplab}                                 & 90.9                 &  & {--}                     & 29.8                & 30.7G                               & 15.4M                                \\
SegFormer$^{\ast}$ \cite{segformer} 				   & 92.3                 &  & {--}                     & 31.7                & 12.1G                               & 7.2M	                           \\
R-CNNP \cite{yolop}                                      & 90.2                 &  & {--}                     & 24.0                & {--}                                & {--}                             	   \\
YOLOP \cite{yolop}                                    & 91.6                 &  & {--}                     & 26.5                & 8.11G                               & 5.53M                                \\
IALaneNet${_{\footnotesize{\text{ResNet-18}}}}$ \cite{interactive}     		   & 90.5                &  & {--}                     & 30.4               & 89.83G                              & 17.05M                               \\
IALaneNet${_{\footnotesize{\text{ResNet-34}}}}$ \cite{interactive}     		   & 90.6                &  & {--}                     & 30.5               & 139.46G                             & 27.16M                               \\
IALaneNet${_{\footnotesize{\text{ConvNeXt-tiny}}}}$ \cite{interactive} 		   & 91.3                &  & {--}                     & 31.5               & 96.52G                              & 18.35M                               \\
IALaneNet${_{\footnotesize{\text{ConvNeXt-small}}}}$ \cite{interactive}  & 91.7                                  &  & {--}                     & \underline{32.5}               & 200.07G                             & 39.97M                               \\
YOLOv8 (multi) \cite{yoloveight}                           & 84.2                 &  & 81.7                     & 24.3                & {--}                                & {--}                                 \\
Sparse U-PDP \cite{sparse} \textcolor{gray}                                & 91.5                 &  & {--}                     & 31.2                & {--}                                & {--}                                 \\
BILane \cite{BILane}                                   & 91.2                  &  & {--}                     & 31.3                 & {--}                                & 1.4M                                \\
TwinLiteNet \cite{twin}                                   & 91.3                 &  & 77.8                     & 31.1                & 3.9G                                & 0.44M                                \\\midrule
\rowcolor{gray3} TwinLiteNet$^+_{\text{Nano}}$                                                    & 87.3                 &  & 70.2                     & 23.3                & \textbf{0.57G}                      & \textbf{0.03M}                                 \\
\rowcolor{gray3} TwinLiteNet$^+_{\text{Small}}$                                                   & 90.6                 &  & 75.8                     & 29.3                & \underline{1.40G}                                & \underline{0.12M}                                \\
\rowcolor{gray3} TwinLiteNet$^+_{\text{Medium}}$                                                  & \underline{92.0}                 &  & \underline{79.1}                     & 32.3                & 4.63G                               & 0.48M                                \\
\rowcolor{gray3} TwinLiteNet$^+_{\text{Large}}$                                                   & \textbf{92.9}        &  & \textbf{81.9}            & \textbf{34.2}       & 17.58G                              & 1.94M                                \\ \bottomrule
\end{tabular}
}
\end{table*}

\subsubsection{\textbf{Drivable Area Segmentation \& Lane Detection Result}}
We evaluate the performance of our TwinLiteNet$^+$ models versus state-of-the-art multitask models that jointly address drivable area segmentation and lane segmentation. For a fair comparison, we exclude single-task models and those incorporating additional tasks such as object detection. Table~\ref{tab: dall} summarizes the results along with computational complexity (FLOPs) and model size (number of parameters).

The original TwinLiteNet achieves 91.3\% mIoU and 31.1\% IoU with only 3.9G FLOPs and 0.44M parameters, offering a strong efficiency–accuracy trade-off among lightweight models. This establishes it as a solid baseline in terms of performance and model compactness. Compared to this baseline, the results demonstrate that TwinLiteNet$^+$ consistently outperforms other lightweight multitask models across all metrics. In particular, the TwinLiteNet$^+_{\text{Large}}$ configuration achieves the highest scores in both tasks-92.9\% mIoU for drivable area segmentation and 34.2\% IoU for lane detection-outperforming strong baselines such as SegFormer, Sparse U-PDP, and IALaneNet. Importantly, TwinLiteNet$^+_{\text{Large}}$ accomplishes this with a significantly smaller model size (1.94M parameters) and moderate FLOPs (17.58G) compared to heavyweight counterparts like IALaneNet or DeepLabV3+, making it highly competitive for both accuracy and efficiency. While the TwinLiteNet$^+_{\text{Nano}}$ model exhibits lower accuracy (mIoU of 87.3\%, IoU of 23.3\%), it offers ultra-efficient inference, requiring just 0.57G FLOPs and 0.03M parameters, making it suitable for highly resource-constrained embedded platforms. The Small and Medium configurations offer balanced trade-offs: TwinLiteNet$^+_{\text{Medium}}$ achieves 92.0\% mIoU and 32.3\% IoU with only 0.48M parameters-surpassing most existing models like YOLOP and even the lightweight BILane (which has 1.4M parameters but slightly lower IoU at 31.3\%).

Overall, these results underscore the scalability and practicality of TwinLiteNet$^+$. The model family provides a spectrum of choices-from highly compact and fast variants to larger, high-accuracy models-empowering developers to choose configurations that best match their target deployment environments. The clear performance advantage of TwinLiteNet$^+$ across different configurations confirms its suitability for both high-performance and real-time embedded applications in autonomous driving.

\subsection{\textbf{Driving perception results in different environments}}

The BDD100K dataset offers diverse annotations for weather conditions, times of the day, and traffic scenes, providing a comprehensive foundation for evaluating model robustness under real-world driving scenarios. To this end, we conduct a detailed analysis of TwinLiteNet$^+_{\text{Large}}$ across multiple environmental contexts. As shown in Table~\ref{perception}, TwinLiteNet$^+_{\text{Large}}$ demonstrates stable performance across a wide range of environmental conditions. Notably, while conditions such as nighttime, Dawn/Dusk, and rainy weather are generally considered challenging due to poor lighting or visual degradation, the model still maintains strong segmentation performance (e.g., 92.5\% mIoU for nighttime, 90.2\% for rainy). This is attributed, in part, to the relatively high number of training samples available for these conditions in the BDD100K dataset, which allows the model to learn more generalized representations. In contrast, the performance tends to drop more significantly in rare scenarios, such as parking lots, tunnels, or gas stations, where the dataset provides few samples ($<$ 500 samples for the training set). In these cases, the segmentation performance declines noticeably, suggesting that data scarcity remains a limiting factor in model generalization under extreme conditions. At the same time, despite having only around 130 samples, foggy scenes still achieve competitive performance-likely because their visual characteristics are not significantly different from those of other weather conditions, allowing the model to transfer learned features effectively.

These findings highlight the importance of both data diversity and representation balance in training robust multi-task models for autonomous driving. The qualitative results under different environmental conditions are further analyzed in Sec. \ref{sec_visualization}

\begin{table}[]
\centering
\setlength{\tabcolsep}{18pt}
\caption{Performance of TwinLiteNet$^+_{\text{Large}}$ evaluated across different environmental conditions. The number of samples used in training and validation is also indicated.}
\label{perception}
\begin{tabular}{llcc}

\toprule
\multirow{2}{*}{\small \textbf{Photometric scene}} & \multirow{2}{*}{\small \textbf{Training/Validation}} & \small \textbf{Drivable Area} &  \textbf{Lane}     \\ \cmidrule(r){3-3} \cmidrule(l){4-4} 
               &                         & \textbf{mIoU (\%)} $\uparrow$     & {\textbf{IoU (\%)}} $\uparrow$ \\ \hline
 \multicolumn{4}{l}{\textit{Weather conditions}} \\
 Clear          & 37,344/5,346                   & 93.1           & 33.8     \\
 Overcast       & 8,770/1,239                   & 93.7          & 35.8     \\
 Undefined      & 8,119/1,157                    & 93.2          & 35.4     \\
 Snowy          & 5,549/769                     & 91.6          & 31.6     \\
 Rainy          & 5,070/738                     & 90.2          & 32.0     \\
 Partly Cloudy  & 4,881/738                     & 93.3          & 35.3     \\
 Foggy          & 130/13                      & 92.4          & 29.6     \\ \hline
 \multicolumn{4}{l}{\textit{Times of the day}} \\ 
 Daytime        & 36,728/5,258                   & 93.2          & 35.1     \\
 Night          & 28,108/3,929                   & 92.5          & 32.9     \\
 Dawn/Dusk      & 5,027/778                     & 92.9          & 33.9     \\
 Undefined      & 137/35                      & 86.7          & 29.3     \\ \hline
\multicolumn{4}{l}{\textit{Traffic scenes}} \\ 
 City street    & 43,516/6,112                   & 92.8           & 34.4     \\
 Highway        & 17,379/2,499                   & 93.3          & 33.8     \\
 Residential    & 8,074/1,253                   & 92.8          & 34.4     \\   
 Parking lot    & 377/49                      & 88.1          & 25.1     \\
 Undefined      & 361/53                      & 90.3          & 31.0     \\
 Tunnel         & 129/27                      & 89.8          & 28.9     \\
 Gas station    & 27/7                       & 88.8          & 14.2     \\ \bottomrule
\end{tabular}
\end{table}

\begin{figure*}[!t]
    \centering
    \includegraphics[width=\textwidth]{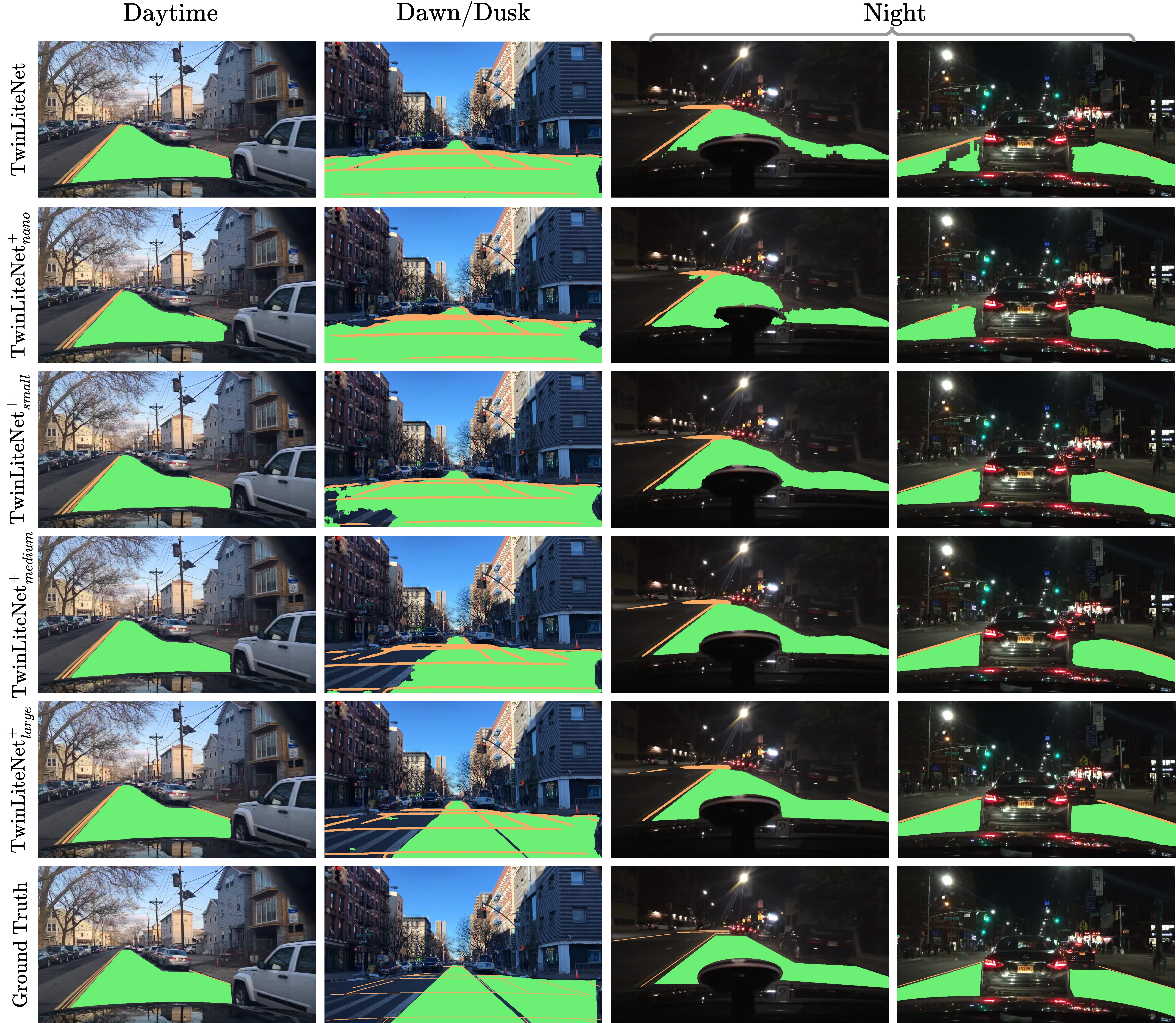}
    \caption{Qualitative results of TwinLiteNet and TwinLiteNet$^+$ at different times of the day.}
    \label{visual_time}
\end{figure*}

\begin{figure*}[!t]
    \centering
    \includegraphics[width=\textwidth]{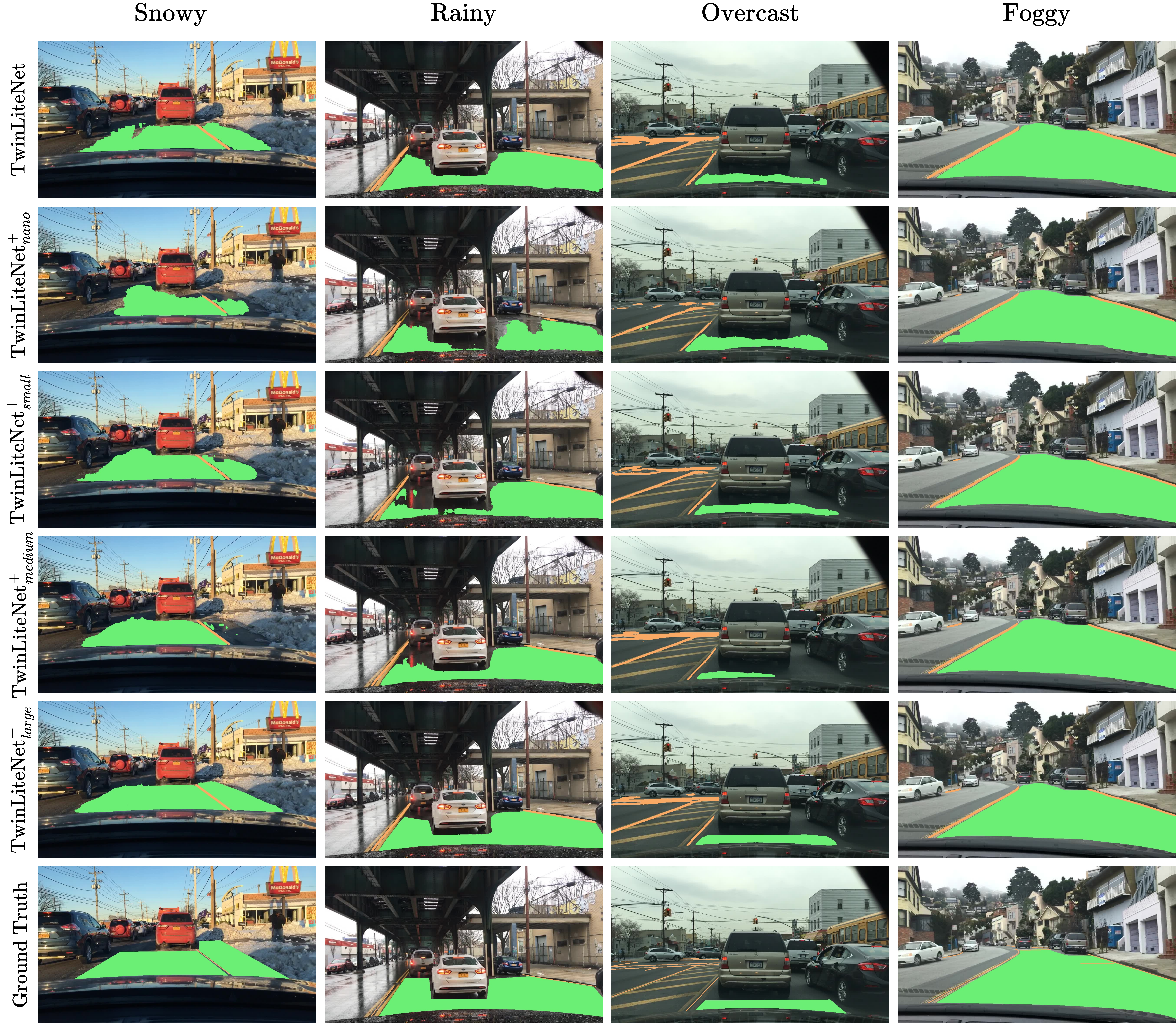}
    \caption{Qualitative results of TwinLiteNet and TwinLiteNet$^+$ in different weather condition.}
    \label{visual_weather}
\end{figure*}

\begin{figure*}[!t]
    \centering
    \includegraphics[width=\textwidth]{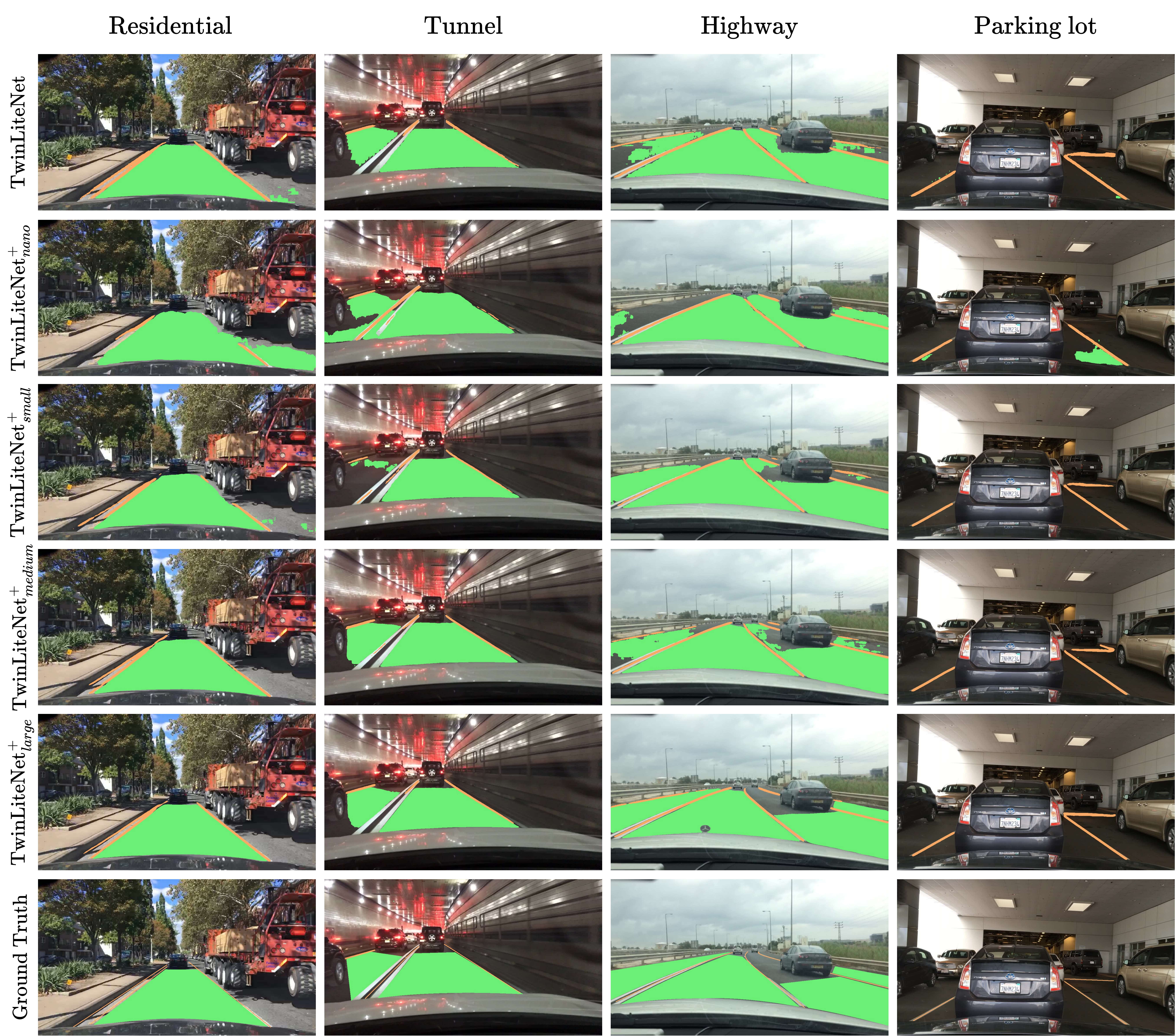}
    \caption{Qualitative results of TwinLiteNet and TwinLiteNet$^+$ in different scene condition.}
    \label{visual_scene}
\end{figure*}

\subsection{\textbf{Visualization}}\label{sec_visualization}

To provide a comprehensive evaluation, we conduct qualitative comparisons of different TwinLiteNet$^+$ configurations against the baseline TwinLiteNet under various environmental conditions. Specifically, we visualize segmentation outputs under (1) different times of the day-including daytime, dawn/dusk, and nighttime-as shown in Figure \ref{visual_time}; (2) Figure \ref{visual_weather} illustrates the results across different weather conditions, including snowy, rainy, overcast, and foggy. and (3) Figure \ref{visual_scene} shows the results under different traffic scenes, including residential areas, tunnels, highways, and parking lots. As shown in the figures, TwinLiteNet$^+_{\text{Large}}$ consistently delivers the most accurate segmentation results, followed by TwinLiteNet$^+_{\text{Medium}}$, which achieves significant improvements over the baseline TwinLiteNet while maintaining similar computational cost. The enhanced performance of TwinLiteNet$^+$ is attributed not only to the integration of the Partial Class Activation Attention (PCAA) module-which facilitates more detailed spatial feature extraction through localized class-specific activations derived from Partial CAM-but also to the strengthened decoder design. The improved decoder incorporates additional convolutional layers after upsampling, allowing for better reconstruction of fine-grained features and contributing to smoother and more precise boundary delineation in segmentation outputs.

Furthermore, the qualitative results reveal that the baseline TwinLiteNet, as well as the lightweight configurations TwinLiteNet$^+_{\text{Nano}}$ and $^+_{\text{Small}}$, encounter challenges in complex environments. These include segmentation noise in cluttered scenes (e.g., residential, parking lot), inaccurate region segmentation (e.g., dawn/dusk), and missed detections of the small lanes (e.g., foggy weather). In contrast, the larger configurations-TwinLiteNet$^+_{\text{Medium}}$ and TwinLiteNet$^+_{\text{Large}}$-demonstrate superior performance in these scenarios due to their greater representational capacity and deeper architecture.

While TwinLiteNet$^+_{\text{Nano}}$ and TwinLiteNet$^+_{\text{Small}}$ demonstrate lower segmentation accuracy, their lightweight architectures provide notable advantages in terms of computational efficiency and inference speed, making them highly suitable for resource-constrained embedded platforms. In contrast, higher-capacity configurations like TwinLiteNet$^+_{\text{Medium}}$ and TwinLiteNet$^+_{\text{Large}}$ consistently achieve more accurate and fine-grained segmentation results, particularly in the aforementioned challenging scenarios, due to their enhanced spatial representation capabilities and deeper model capacity. These observations highlight the flexibility and scalability of the TwinLiteNet$^+$ configs, enabling practical deployment across a wide range of autonomous driving applications-from high-performance systems to energy-efficient real-time use cases.

\subsection{\textbf{Practical Deployment Considerations}} \label{deployment}

\begin{table*}[!t]
\centering
\caption{Performance benchmarking of TwinLiteNet+ across different Quantization approaches: Floating Point 16-bit (FP16) and Integer 8-bit (INT8). $\downarrow$} indicates a performance drop caused by quantization, whereas \textcolor{gray}{$\rightarrow$} signifies that the accuracy remains unchanged.
\label{8bit}
\resizebox{\columnwidth}{!}{
\begin{tabular}{lcccclcccclcccc}
\toprule
\multirow{4}{*}{\textbf{Config}}               & \multicolumn{4}{c}{\textbf{Drivable Area}}                                                                                                                                                                                               &  & \multicolumn{9}{c}{\textbf{Lane}}                                                                                                                                                                                                                                                                           \\ \cmidrule{2-5} \cmidrule{7-15} 
                                & \multicolumn{4}{c}{\textbf{mIoU (\%)}$\uparrow$}                                                                                                                                                                                                   &  & \multicolumn{4}{c}{\textbf{Acc (\%)}$\uparrow$}                                                                                                                                                                                                                                &  & \multicolumn{4}{c}{\textbf{IoU (\%)}$\uparrow$}                                                                                                                                                                                                                                \\ \cmidrule{2-5} \cmidrule{7-10} \cmidrule{12-15} 
                                & \multirow{2}{*}{\textbf{FP32}} & \multirow{2}{*}{\textbf{FP16}}                    & \multicolumn{2}{c}{\textbf{INT8}}                                            &  & \multirow{2}{*}{\textbf{FP32}} & \multirow{2}{*}{\textbf{FP16}}                    & \multicolumn{2}{c}{\textbf{INT8}}                                            &  & \multirow{2}{*}{\textbf{FP32}}          & \multirow{2}{*}{\textbf{FP16}}                    & \multicolumn{2}{c}{\textbf{INT8}}                                            \\
                                &                       &                                                 & \textbf{PTSQ}                        & \textbf{QAT}                         &  &                       &                                                 & \textbf{PTSQ}                        & \textbf{QAT}                         &  &               &                                                 & \textbf{PTSQ}                        & \textbf{QAT}                         \\ \midrule
Nano   & \textbf{87.3}         & \begin{tabular}[c]{@{}c@{}}87.3\\ \textcolor{gray}{$\rightarrow$ 0.0}\end{tabular} & \begin{tabular}[c]{@{}c@{}}86.0\\ {\color[HTML]{FE0000} $\downarrow$ 1.3}\end{tabular} & \begin{tabular}[c]{@{}c@{}}86.9\\ {\color[HTML]{FE0000} $\downarrow$ 0.4}\end{tabular} &  & \textbf{70.2}         & \begin{tabular}[c]{@{}c@{}}70.1\\ {\color[HTML]{FE0000} $\downarrow$ 0.1}\end{tabular} & \begin{tabular}[c]{@{}c@{}}67.2\\ {\color[HTML]{FE0000} $\downarrow$ 3.0}\end{tabular} & \begin{tabular}[c]{@{}c@{}}69.9\\ {\color[HTML]{FE0000} $\downarrow$ 0.3}\end{tabular} &  & \textbf{23.3} & \begin{tabular}[c]{@{}c@{}}23.2\\ {\color[HTML]{FE0000} $\downarrow$ 0.1}\end{tabular} & \begin{tabular}[c]{@{}c@{}}21.8\\ {\color[HTML]{FE0000} $\downarrow$ 1.5}\end{tabular} & \begin{tabular}[c]{@{}c@{}}23.2\\ {\color[HTML]{FE0000} $\downarrow$ 0.1}\end{tabular} \\
Small  & \textbf{90.6}         & \begin{tabular}[c]{@{}c@{}}90.6\\ \textcolor{gray}{$\rightarrow$ 0.0}\end{tabular} & \begin{tabular}[c]{@{}c@{}}88.5\\ {\color[HTML]{FE0000} $\downarrow$ 2.1}\end{tabular} & \begin{tabular}[c]{@{}c@{}}90.2\\ {\color[HTML]{FE0000} $\downarrow$ 0.4}\end{tabular} &  & \textbf{75.8}         & \begin{tabular}[c]{@{}c@{}}75.7\\ {\color[HTML]{FE0000} $\downarrow$ 0.1}\end{tabular} & \begin{tabular}[c]{@{}c@{}}74.6\\ {\color[HTML]{FE0000} $\downarrow$ 1.2}\end{tabular} & \begin{tabular}[c]{@{}c@{}}75.6\\ {\color[HTML]{FE0000} $\downarrow$ 0.2}\end{tabular} &  & \textbf{29.3} & \begin{tabular}[c]{@{}c@{}}29.3\\ \textcolor{gray}{$\rightarrow$ 0.0}\end{tabular} & \begin{tabular}[c]{@{}c@{}}28.3\\ {\color[HTML]{FE0000} $\downarrow$ 1.0}\end{tabular} & \begin{tabular}[c]{@{}c@{}}28.8\\ {\color[HTML]{FE0000} $\downarrow$ 0.5}\end{tabular} \\
Medium & \textbf{92.0}         & \begin{tabular}[c]{@{}c@{}}92.0\\ \textcolor{gray}{$\rightarrow$ 0.0}\end{tabular} & \begin{tabular}[c]{@{}c@{}}91.6\\ {\color[HTML]{FE0000} $\downarrow$ 0.4}\end{tabular} & \begin{tabular}[c]{@{}c@{}}91.9\\ {\color[HTML]{FE0000} $\downarrow$ 0.1}\end{tabular} &  & \textbf{79.1}         & \begin{tabular}[c]{@{}c@{}}79.1\\ \textcolor{gray}{$\rightarrow$ 0.0}\end{tabular} & \begin{tabular}[c]{@{}c@{}}77.1\\ {\color[HTML]{FE0000} $\downarrow$ 2.0}\end{tabular} & \begin{tabular}[c]{@{}c@{}}78.6\\ {\color[HTML]{FE0000} $\downarrow$ 0.5}\end{tabular} &  & \textbf{32.3} & \begin{tabular}[c]{@{}c@{}}32.2\\ {\color[HTML]{FE0000} $\downarrow$ 0.1}\end{tabular} & \begin{tabular}[c]{@{}c@{}}31.5\\ {\color[HTML]{FE0000} $\downarrow$ 0.8}\end{tabular} & \begin{tabular}[c]{@{}c@{}}31.9\\ {\color[HTML]{FE0000} $\downarrow$ 0.4}\end{tabular} \\
Large  & \textbf{92.9}         & \begin{tabular}[c]{@{}c@{}}92.9\\ \textcolor{gray}{$\rightarrow$ 0.0}\end{tabular} & \begin{tabular}[c]{@{}c@{}}92.7\\ {\color[HTML]{FE0000} $\downarrow$ 0.2}\end{tabular} & \begin{tabular}[c]{@{}c@{}}92.7\\ {\color[HTML]{FE0000} $\downarrow$ 0.2}\end{tabular} &  & \textbf{81.9}         & \begin{tabular}[c]{@{}c@{}}81.9\\ \textcolor{gray}{$\rightarrow$ 0.0}\end{tabular}& \begin{tabular}[c]{@{}c@{}}80.2\\ {\color[HTML]{FE0000} $\downarrow$ 1.7}\end{tabular} & \begin{tabular}[c]{@{}c@{}}81.4\\ {\color[HTML]{FE0000} $\downarrow$ 0.5}\end{tabular} &  & \textbf{34.2} & \begin{tabular}[c]{@{}c@{}}34.2\\ \textcolor{gray}{$\rightarrow$ 0.0}\end{tabular} & \begin{tabular}[c]{@{}c@{}}33.6\\ {\color[HTML]{FE0000} $\downarrow$ 0.6}\end{tabular} & \begin{tabular}[c]{@{}c@{}}34.0\\ {\color[HTML]{FE0000} $\downarrow$ 0.2}\end{tabular} \\ \bottomrule
\end{tabular}
}
\end{table*}

\begin{table*}[!b]
\centering
\setlength{\tabcolsep}{10pt}
\caption{Latency and Power Consumption benchmarking of TwinLiteNet$^+$ on Embedded Devices (Jetson Xavier and Jetson TX2) in different Quantization methods.}
\label{tab:latency}
\begin{tabular}{lccclcc}
\toprule
\multicolumn{1}{l}{\multirow{2}{*}{\textbf{Config}}} & \multicolumn{3}{c}{\textbf{Jetson Xavier}}                         &  & \multicolumn{2}{c}{\textbf{Jetson TX2}}     \\ \cmidrule{2-4} \cmidrule{6-7} 
\multicolumn{1}{l}{}  & \textbf{INT8}                 & \textbf{FP16}                 & \textbf{FP32}                 &  & \textbf{FP16}                 & \textbf{FP32}                 \\ \midrule
 \multicolumn{4}{l}{\textit{Latency (ms) $\downarrow$}} \\ 
 Nano                                                 & \textcolor{white}{0}7.553$_{\pm 0.046}$    & \textcolor{white}{0}9.155$_{\pm 0.092}$   & 10.303$_{\pm 0.222}$   &  & \textcolor{white}{0}22.268$_{\pm 0.156}$   & \textcolor{white}{0}26.374$_{\pm 0.058}$   \\
Small                                                & 10.286$_{\pm 0.093}$   & 12.861$_{\pm 0.941}$   & 15.662$_{\pm 0.055}$   &  & \textcolor{white}{0}33.818$_{\pm 0.134}$   & \textcolor{white}{0}41.961$_{\pm 0.035}$   \\
Medium                                               & 15.548$_{\pm 0.083}$   & 20.460$_{\pm 0.067}$   & 27.985$_{\pm 0.166}$   &  & \textcolor{white}{0}61.785$_{\pm 0.015}$   & \textcolor{white}{0}84.631$_{\pm 0.096}$   \\
Large                                                & 29.102$_{\pm 0.107}$   & 43.089$_{\pm 0.198}$   & 69.150$_{\pm 0.382}$   &  & 153.240$_{\pm 0.094}$   & 218.150$_{\pm 0.099}$  \\

\midrule
 \multicolumn{4}{l}{\textit{Power (Watt) $\downarrow$}} \\ 
Nano                                                 & \textcolor{white}{0}9.657$_{\pm 2.034}$    & 10.029$_{\pm 2.267}$   & 11.790$_{\pm 2.089}$   &  & \textcolor{white}{00}3.647$_{\pm 0.637}$    & \textcolor{white}{00}4.097$_{\pm 0.603}$    \\ 
Small                                                & 10.926$_{\pm 1.924}$   & 11.086$_{\pm 2.385}$   & 12.732$_{\pm 2.505}$   &  & \textcolor{white}{00}4.043$_{\pm 0.463}$    & \textcolor{white}{00}4.513$_{\pm 0.376}$    \\
Medium                                               & 12.427$_{\pm 2.086}$   & 13.585$_{\pm 1.832}$   & 15.496$_{\pm 1.549}$   &  & \textcolor{white}{00}4.496$_{\pm 0.202}$    & \textcolor{white}{00}4.747$_{\pm 0.122}$    \\
Large                                                & 14.666$_{\pm 1.632}$   & 15.980$_{\pm 1.187}$   & 17.710$_{\pm 0.594}$   &  & \textcolor{white}{00}4.883$_{\pm 0.057}$    & \textcolor{white}{00}5.019$_{\pm 0.051}$    \\ \bottomrule
\end{tabular}
\end{table*}

\subsubsection{Quantization} \label{quantization_sec}

To evaluate the practical deployability of the TwinLiteNet$^+$ model on hardware with limited computational resources, we conducted experiments using three common inference data types: 32-bit floating point (FP32), 16-bit floating point (FP16), and 8-bit integer (INT8). Among these, FP16 can be applied directly without requiring retraining or calibration. In contrast, quantization to INT8 demands additional processing to preserve accuracy.

Specifically, for Post-Training Static Quantization (PTSQ), we followed a standard calibration procedure by sampling 500 random images from the training set to estimate activation ranges. Based on this, we applied 8-bit quantization to both weights and activations. While this approach is straightforward to integrate post-training, it resulted in reduced segmentation accuracy, particularly in lane detection, where pixel-level precision is crucial. To mitigate this problem, we employed Quantization-Aware Training (QAT), which simulates quantization effects during training. Using the \texttt{pytorch\_quantization} framework, we inserted fake quantization modules into the model's computational graph. Rather than training from scratch, we fine-tuned a pre-trained model for an additional 10 epochs, enabling it to adapt to quantization noise while maintaining training stability.

As shown in Table~\ref{8bit}, TwinLiteNet$^+$ maintains strong segmentation performance under all quantization settings. While INT8 quantization involves a more complex process than FP16, QAT allows the model to retain accuracy comparable to FP32 while significantly improving inference efficiency and reducing model size-critical factors for deployment on resource-constrained embedded devices.

\subsubsection{Inference Performance on Embedded Devices}

Furthermore, we implement our model on several embedded devices such as Jetson Xavier and Jetson TX2. TensorRT-FP32, FP16 are used respectively for inference and then measured the inference latency and power consumption on each device. With TensorRT-INT8, only Jetson Xavier is measured, as the Jetson TX2 isn't supported yet. To enable INT8 quantization, we adopt the Post-Training Static Quantization (PTSQ) method, as our primary focus is to assess inference efficiency rather than optimize quantization accuracy. In this process, we perform calibration using 500 randomly selected samples from the training set to estimate the activation ranges, following the setup described in Section~\ref{quantization_sec}. All inference experiments are performed using TensorRT with default settings, without any manual tuning or hardware-specific optimization. This setup aims to evaluate how well the model performs on various embedded devices. To minimize measurement discrepancies, we conduct five iterations, with each iteration performing 100 inferences. Results from Table \ref{tab:latency} show that using the TwinLiteNet$^+$ model on embedded devices like Jetson Xavier and Jetson TX2 yields promising results in terms of inference latency and power consumption. Specifically, the Nano and Small configs of the model exhibited low latency and reasonable power consumption on both devices with TensorRT-INT8, highlighting the model's performance optimization capability in a limited computational environment. Although the Medium and Large configs have higher latency and power consumption, they still maintain an acceptable level, demonstrating the model's flexibility and scalability. These results not only validate the impressive performance of TwinLiteNet$^+$ but also underscore its practical potential in real-world applications, particularly in the field of autonomous vehicles and smart embedded systems.

\subsection{\textbf{Directly $\&$ Alternative Area segmentation results}}

\begin{figure}[!t]
  \centering
    \begin{subfigure}{0.32\columnwidth}
        \includegraphics[width=\textwidth]{al1.jpg}
        \caption{Ground truth\\
        {\color[HTML]{FFFFFF} .}\\
        {\color[HTML]{FFFFFF} .}
        } 
    \end{subfigure} 
    \begin{subfigure}{0.32\columnwidth}
        \includegraphics[width=\textwidth]{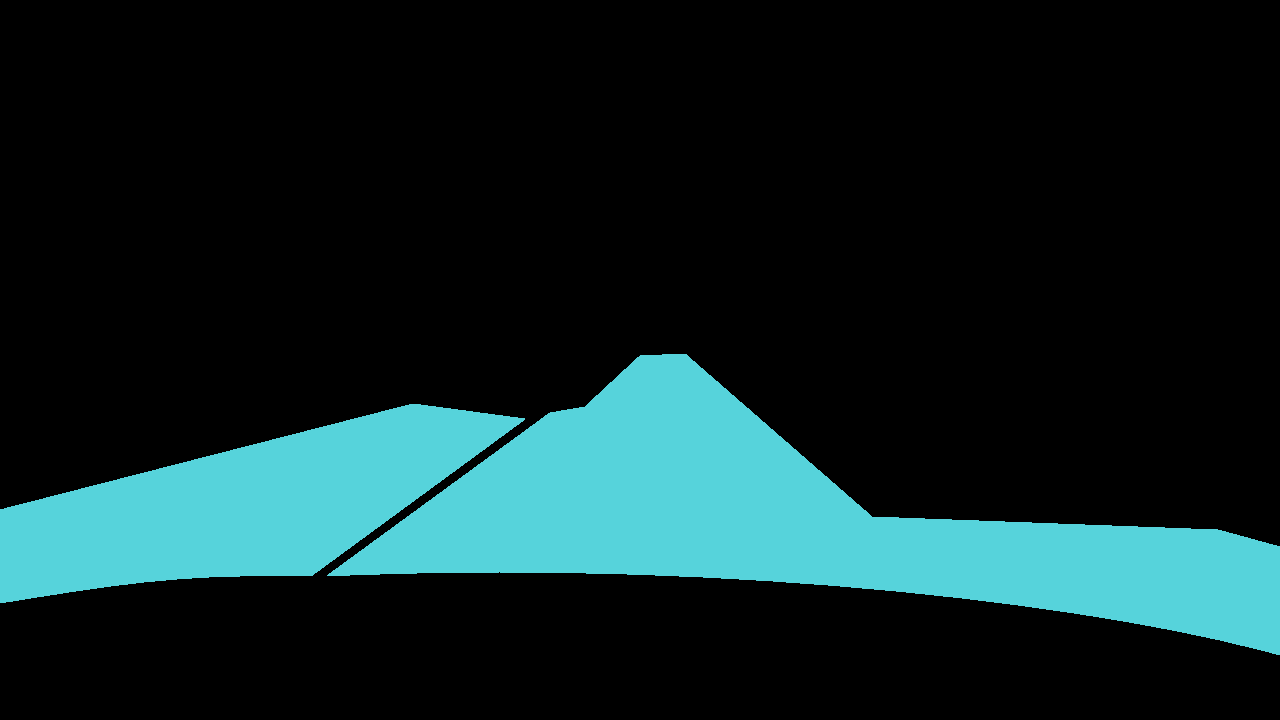}
        \caption{Drivable Area Segmentation\\
        {\color[HTML]{FFFFFF} .}\\
        {\color[HTML]{FFFFFF} .}
        \label{6b}
        } 
    \end{subfigure} 
    \begin{subfigure}{0.32\columnwidth}
        \includegraphics[width=\textwidth]{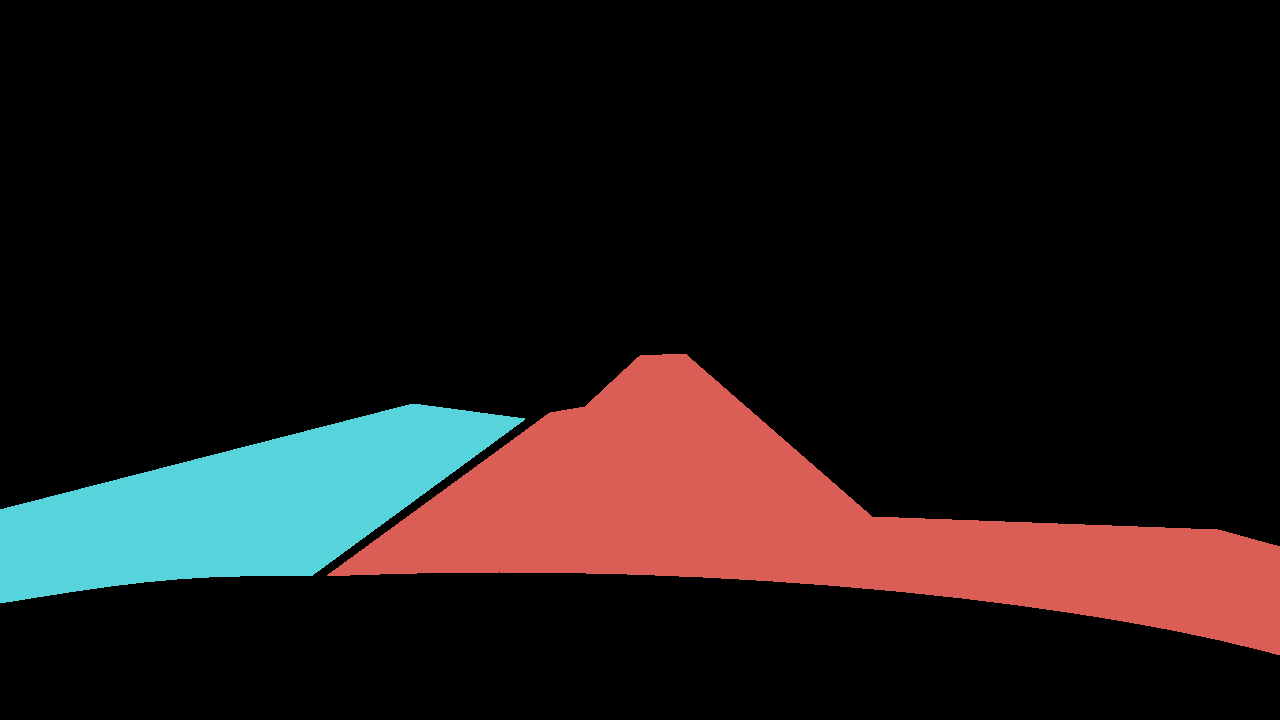}
        \caption{Directly \& Alternative Area segmentation\\
        \label{6c}
        } 
    \end{subfigure} 
    \caption{Examples of Ground truth visualization for Directly \& Alternative Area segmentation task.}
    \label{alter}
\end{figure}

In this section, we expand our research to include a model that focuses on the division of directly drivable and alternative areas. This approach differs from the standard drivable area segmentation, which usually combines these two distinct areas into a single area. Figure \ref{alter} illustrates the main difference between the two segmentation tasks, depicted in Figure \ref{6b} and Figure \ref{6c}, where Figure \ref{6c} is segmented into direct and alternative areas. This division of regions allows vehicles to differentiate between navigable regions and alternative routes, making it easier to change lanes or maneuver around obstacles in the region. When performing this segmentation, the controllable segmentation decoder block output has three channels instead of two. Therefore, the output of the model for the driving area segment is $\textbf{O}_{drivable} \in \mathbb{R}^{3 \times H \times W}$. We train the model using two designs: One with a single decoder for direct and alternative driving area segmentation and another with two decoders for multi-task models (Lane segmentation and Directly $\&$ Alternative Area segmentation).

\begin{table}[!t]
\caption{Perforance benchmarking in mIoU (\%), PA (\%) and mPA (\%) between different models in Directly \& Alternative Area segmentation tasks. The best and second best results are marked in \textbf{bold} and \underline{underline} respectively.}
\centering
\setlength{\tabcolsep}{22pt}
% \resizebox{1.0\columnwidth}{!}{
\begin{tabular}{lcccc}
\toprule
\multirow{2}{*}{\textbf{Model}}                                                 & \multicolumn{3}{c}{\textbf{Directly \& Alternative Area Segmentation}}              \\ \cmidrule{2-4} 
                                                                       & \textbf{mIoU (\%)} $\uparrow$ & \textbf{PA(\%)} $\uparrow$ & \textbf{mPA(\%)} $\uparrow$ \\ \midrule
DeepLabv3+ \cite{deeplab}              & \textbf{84.73}               & --                & --                 \\
PSPNet \cite{pspnet}      & 84.13               & --                & --                 \\
\textit{Pizzati et al.} \cite{enhanced}                        & 82.62                & --                & --                 \\
ShuDA-RFBNet \cite{shuda}             & 82.67                & --                & --                 \\

Resnet+ASPP+FPN \cite{aspp}       & \underline{84.58}                & \textbf{97.09}    & 91.14              \\
Resnet+FPN \cite{aspp}           & 82.70                & 96.51             & 91.42              \\
Resnet+ASPP+top-down \cite{aspp} & 82.80                 & 96.60             & 90.68              \\
Resnet+top-down \cite{aspp}      & 82.44                & 96.24             & 88.39              \\
\textit{Lee} \cite{fast}                          & 83.34                & --                & --                 \\
\textit{Sun et al.} \cite{lightweight}                 & 83.01                & --                & --                 \\
IDS-MODEL \cite{idsmodel}            & 83.63                & --                & --                 \\ \midrule
\rowcolor{gray3} TwinLiteNet$^+_{\text{\tiny{D\&A}}}$ \footnotesize{(single-task)}                            & 83.9                 & 96.9              & \underline{92.0}               \\
\rowcolor{gray3} TwinLiteNet$^+_{\text{\tiny{D\&A}}}$ \footnotesize{(multi-task)}                             & 84.3                 & \underline{97.0}              & \textbf{92.1}      \\ \bottomrule
\end{tabular}%
% }
\label{tab: 2dall}
\end{table}

\begin{figure}[!t]
\centering
\includegraphics[width=\linewidth]{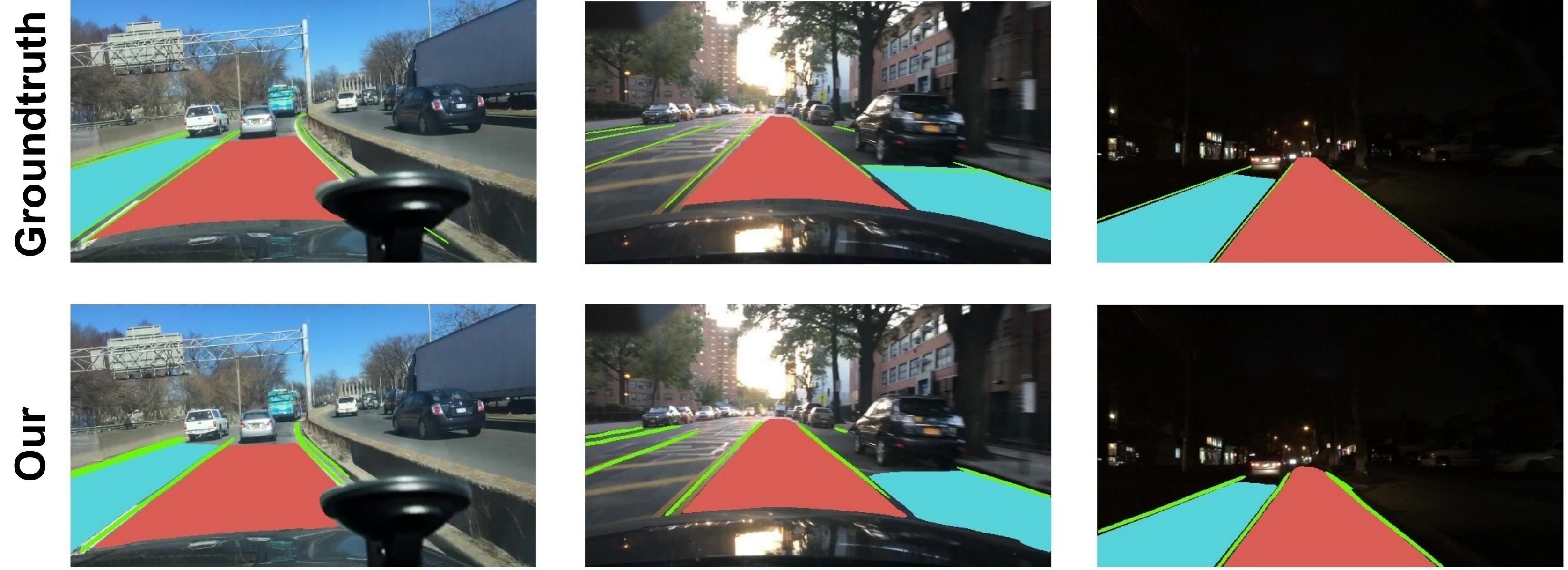}
\caption{Results visualization of TwinLiteNet$^+_{\text{D\&A}}$ for Directly $\&$ Alternative Area segmentation. {\color[HTML]{FD6864}Red} regions are directly drivable area, the {\color[HTML]{38FFF8}blue} ones are alternative and the lanes are {\color[HTML]{34FF34}green}.}    
\label{alterre}
\end{figure}

Table \ref{tab: 2dall} provides a comprehensive and detailed overview of the performance of various segmentation models applied to direct region segmentation and replacement. In addition to the evaluation mIoU metric, in this study, we also assess the Pixel Accuracy (PA) and Mean Pixel Accuracy (mPA), facilitating comparison with previous research. In this experiment, we compare the TwinLiteNet$^+_{\text{Large}}$ model after modifications to demonstrate the responsiveness of our model to this task, and we name it as TwinLiteNet$^+_{\text{D\&A}}$ to differentiate it from the Large one. Additionally, we experiment with the TwinLiteNet$^+_{\text{D\&A}}$ model when performing multi-tasks and the TwinLiteNet$^+_{\text{D\&A}}$ with only one decoding block for the driving area segmentation. The TwinLiteNet$^+_{\text{D\&A}}$ models, in both multitask and single-task configurations, stand out with high performance across all three metrics, with our TwinLiteNet$^+_{\text{D\&A}}$ model achieving the highest mPA of 92.1\%, surpassing previous studies and competitive in other metrics. This comprehensive assessment underscores the advancements in segmentation techniques and highlights the TwinLiteNet$^+_{\text{D\&A}}$ model's potential to improve accuracy and detail capture in image segmentation tasks. We present some results of the TwinLiteNet$^+_{\text{D\&A}}$ model in Figure \ref{alterre} to emphasize its proficiency in differentiating between directly drivable and alternative paths under various lighting conditions, from daylight to nighttime scenarios.

% \usepackage{multirow}

\subsection{\textbf{Ablation study}} \label{ablation}

\subsubsection{Multi-task versus single-task} \label{ablation_stvsmt}

To evaluate the effectiveness of the multitask learning strategy, we compare the performance of the TwinLiteNet$^+_{\text{Large}}$ model in multitask and single-task configurations. In the single-task settings, we retain only the relevant decoder-either for drivable area or lane segmentation-while removing the other to isolate performance. The results are summarized in Table~\ref{ab1}.

For the drivable area segmentation task, the multitask version slightly outperforms its single-task counterpart, achieving a 0.3\% gain in mIoU. This improvement suggests that shared representations with the lane segmentation task positively influence the learning process by providing complementary structural cues. For the lane segmentation task, multitask learning maintains the same lane accuracy (81.9\%) but experiences a slight drop of 0.2\% in IoU, which may result from representational interference between the tasks. These findings indicate that while multitask learning introduces minor trade-offs-particularly for tasks sensitive to spatial precision-it also offers measurable advantages in feature sharing and overall model compactness.

In terms of efficiency, the multitask model incurs only a marginal increase in both parameter count (+0.03M) and computational cost (+0.61 GFLOPs) compared to the single-task variants. This slight overhead primarily stems from the addition of a single extra decoder to support the second task, while the encoder is shared between both. In contrast, deploying two separate single-task models would require duplicating the entire backbone, leading to a significantly higher computational cost and memory footprint. Thus, the multitask approach not only enables joint optimization and feature sharing but also offers a more resource-efficient alternative to running independent models. These findings reinforce the practical value of multitask learning for real-time autonomous driving applications, where both accuracy and efficiency are critical.

\begin{table*}[t]
\centering
\caption{TwinLiteNet$^+$ model's evaluation between Multitasking Learning versus Single-Task Learning approach. The best results are marked in \textbf{bold}. $\uparrow$ indicates increasing, $\downarrow$ indicates decreasing, and $\rightarrow$ represents no change. \textcolor{red}{Red} denotes negative impact, while \textcolor{blue}{blue} highlights gains achieved through multi-task training.}
\label{ab1}
\resizebox{1\columnwidth}{!}{
\begin{tabular}{lclcccc}
\toprule
\multirow{2}{*}{\textbf{Method}} & \textbf{Drivable Area}                                                                        &  & \multicolumn{2}{c}{\textbf{Lane}}                                                                                                                                                        & \multirow{2}{*}{\textbf{Parameter} $\downarrow$}                                                 & \multirow{2}{*}{\textbf{FLOPs} $\downarrow$}                                                   \\ \cmidrule{2-2} \cmidrule{4-5}
                        & \textbf{mIoU (\%)} $\uparrow$                                                                 &  & \textbf{Acc (\%)} $\uparrow$                                                                    & \textbf{IoU (\%)} $\uparrow$                                                                    &                                                                                         &                                                                                        \\ \midrule
Drivable Area (only)         & 92.6                                                                                 &  & \xmark                                                                                     & \xmark                                                                                     & \textbf{1.91M}                                                                          & \textbf{16.97G}                                                                         \\
Lane (only)             & \xmark                                                                                   &  & \textbf{81.9}                                                                          & \textbf{34.4}                                                                          & \textbf{1.91M}                                                                          & \textbf{16.97G}                                                                         \\ \midrule
 \rowcolor{gray3} Multi-task              & \begin{tabular}[c]{@{}c@{}}\textbf{92.9}\\ {\color[HTML]{3531FF} $\uparrow$ 0.3}\end{tabular} &  & \begin{tabular}[c]{@{}c@{}}\textbf{81.9}\\ \textcolor{gray}{$\rightarrow$ 0.0}\end{tabular} & \begin{tabular}[c]{@{}c@{}}34.2\\ {\color[HTML]{FE0000} $\downarrow$ 0.2}\end{tabular} & \begin{tabular}[c]{@{}c@{}}1.94M \\ {\color[HTML]{FE0000} $\uparrow$ 0.03}\end{tabular} & \begin{tabular}[c]{@{}c@{}}17.58G\\ {\color[HTML]{FE0000} $\uparrow$ 0.61}\end{tabular} \\ \bottomrule
\end{tabular}
}
\end{table*}

\subsubsection{The ablations in different experimental settings}

To better understand the contribution of each component in the proposed TwinLiteNet$^+$ architecture, we perform an ablation study based on the TwinLiteNet$^+_{\text{Large}}$ configuration. Each experiment isolates a specific module to evaluate its impact on segmentation performance and model complexity, with quantitative results summarized in Table~\ref{ab2}. Six modified variants are constructed as follows: (1) \textit{Tversky Loss:} removed by replacing the combined objective with Focal Loss alone for both segmentation branches; (2) \textit{DESP:} replaced with the original ESP module from the baseline TwinLiteNet; (3) \textit{Robust Decoder:} disabled by reverting to the simple decoder design of the TwinLiteNet baseline, thus not use the UCB and USB upsampling proposed modules; (4) \textit{Multiple Decoder:} replaced with a shared single decoder head producing a unified three-channel output $\mathbf{O} \in \mathbb{R}^{3 \times H \times W}$ for both tasks; (5) \textit{PCAA Module:} removed by bypassing the attention mechanism and directly linking the encoder output to the decoder; and (6) \textit{EMA (Exponential Moving Average):} deactivated by using the model’s raw weights instead of the smoothed exponential averages accumulated during training.

Experimental findings show that each component plays an important role in the model’s effectiveness. The removal of Tversky Loss results in a few performance drops in drivable area segmentation. However, it is significant in lane segmentation, highlighting its crucial role in managing class imbalance and improving recall for thin and sparse structures. Substituting DESP with the baseline ESP architecture slightly increases lane segmentation accuracy. However, it introduces a notable rise in FLOPs and parameter count, underscoring the importance of depthwise separable convolution in achieving efficiency. Eliminating any of the Robust Decoder, Multiple Decoder, or PCAA modules leads to consistent degradation across both drivable area and lane segmentation metrics, confirming their importance in enhancing spatial detail, semantic refinement, and task-specific decoding. Lastly, omitting EMA leads to minor reductions in performance, suggesting its stabilizing influence on model convergence and generalization.

The results from the ablation experiments demonstrate that each component in the design of TwinLiteNet$^+$ plays a vital role-not only enhancing accuracy in both segmentation tasks but also optimizing computational cost.

\begin{table*}[!t]
\centering
\setlength{\tabcolsep}{5pt}
\caption{TwinLiteNet$^+_{Large}$ Model's evaluation with different experimental settings. Across different variants of TwinLiteNet$^+$, the symbols $\uparrow$, $\rightarrow$ and $\downarrow$ indicate changes in both performance metrics and computational cost. Specifically, $\uparrow$ represents an improvement in accuracy (mIoU, Acc, IoU) or an increase in computational overhead, while $\downarrow$ indicates a reduction. Additionally, \textcolor{red}{red-colored} values highlight negative changes, whereas \textcolor{blue}{blue-colored} values indicate positive improvements.}
\label{ab2}
% \resizebox{1\columnwidth}{!}{
\begin{tabular}{lclcccc}
\toprule
\multicolumn{1}{c}{\multirow{2}{*}{\textbf{}}} & \textbf{Drivable Area}                                                             &  & \multicolumn{2}{c}{\textbf{Lane}}                                                                                                                                     & \multirow{2}{*}{\textbf{FLOPs} $\downarrow$}                                                     & \multirow{2}{*}{\textbf{\#Param. $\downarrow$}}                                                  \\ \cline{2-2} \cline{4-5}
\multicolumn{1}{c}{}                           & \textbf{mIoU (\%) $\uparrow$}                                                                 &  & \textbf{Acc (\%) $\uparrow$}                                                                & \textbf{IoU (\%) $\uparrow$}                                                                  &                                                                                      &                                                                                     \\ \midrule
\rowcolor{gray3} TwinLiteNet$^+_{Large}$    & \textbf{92.9}                                                                               &  & \underline{81.9}                                                                             & \textbf{34.2}                                                                               & 17.58G                                                                      & 1.94M                                                                    \\ \midrule
\textit{\small{w/o Tversky Loss}}                      & \begin{tabular}[c]{@{}c@{}}\underline{92.7}\\ \textcolor{red}{$\downarrow$ 0.2}\end{tabular}   &  & \begin{tabular}[c]{@{}c@{}}79.7\\ \textcolor{red}{$\downarrow$ 2.2}\end{tabular} & \begin{tabular}[c]{@{}c@{}}31.1\\ \textcolor{red}{$\downarrow$ 3.1}\end{tabular}   & \begin{tabular}[c]{@{}c@{}}17.58G\\ \textcolor{gray}{$\rightarrow$ 0.0}\end{tabular}   & \begin{tabular}[c]{@{}c@{}}1.94M\\ \textcolor{gray}{$\rightarrow$ 0.0}\end{tabular}   \\
\textit{\small{w/o DESP proposed}}                              & \begin{tabular}[c]{@{}c@{}}\textbf{92.9}\\ \textcolor{gray}{$\rightarrow$ 0.0}\end{tabular} &  & \begin{tabular}[c]{@{}c@{}}\textbf{82.2}\\ \textcolor{blue}{$\uparrow$ 0.3}\end{tabular}  & \begin{tabular}[c]{@{}c@{}}\textbf{34.2}\\ \textcolor{gray}{$\rightarrow$ 0.0}\end{tabular} & \begin{tabular}[c]{@{}c@{}}22.19G\\ \textcolor{red}{$\uparrow$ 4.61}\end{tabular}    & \begin{tabular}[c]{@{}c@{}}2.78M\\ \textcolor{red}{$\uparrow$ 0.84}\end{tabular}    \\
\textit{\small{w/o Robust Decoder}}                    & \begin{tabular}[c]{@{}c@{}}92.4\\ \textcolor{red}{$\downarrow$ 0.5}\end{tabular}   &  & \begin{tabular}[c]{@{}c@{}}81.1\\ \textcolor{red}{$\downarrow$ 0.8}\end{tabular} & \begin{tabular}[c]{@{}c@{}}33.3\\ \textcolor{red}{$\downarrow$ 0.9}\end{tabular}   & \begin{tabular}[c]{@{}c@{}}\textbf{16.78G}\\ \textcolor{blue}{$\downarrow$ 0.80}\end{tabular} & \begin{tabular}[c]{@{}c@{}}\underline{1.90M}\\ \textcolor{blue}{$\downarrow$ 0.04}\end{tabular} \\
\textit{\small{w/o Multiple Decoder}}                  & \begin{tabular}[c]{@{}c@{}}91.3\\ \textcolor{red}{$\downarrow$ 1.6}\end{tabular}   &  & \begin{tabular}[c]{@{}c@{}}79.9\\ \textcolor{red}{$\downarrow$ 2.0}\end{tabular} & \begin{tabular}[c]{@{}c@{}}32.1\\ \textcolor{red}{$\downarrow$ 2.1}\end{tabular}   & \begin{tabular}[c]{@{}c@{}}\underline{16.99G}\\ \textcolor{blue}{$\downarrow$ 0.59}\end{tabular} & \begin{tabular}[c]{@{}c@{}}1.91M\\ \textcolor{blue}{$\downarrow$ 0.03}\end{tabular} \\
\textit{\small{w/o PCAA Module}}                              & \begin{tabular}[c]{@{}c@{}}92.5\\ \textcolor{red}{$\downarrow$ 0.4}\end{tabular}   &  & \begin{tabular}[c]{@{}c@{}}81.4\\ \textcolor{red}{$\downarrow$ 0.5}\end{tabular} & \begin{tabular}[c]{@{}c@{}}33.8\\ \textcolor{red}{$\downarrow$ 0.4}\end{tabular}   & \begin{tabular}[c]{@{}c@{}}17.42G\\ \textcolor{blue}{$\downarrow$ 0.16}\end{tabular} & \begin{tabular}[c]{@{}c@{}}\textbf{1.82M}\\ \textcolor{blue}{$\downarrow$ 0.12}\end{tabular} \\
\textit{\small{w/o EMA}}                               & \begin{tabular}[c]{@{}c@{}}\underline{92.7}\\ \textcolor{red}{$\downarrow$ 0.2}\end{tabular}   &  & \begin{tabular}[c]{@{}c@{}}81.7\\ \textcolor{red}{$\downarrow$ 0.2}\end{tabular} & \begin{tabular}[c]{@{}c@{}}\underline{33.9}\\ \textcolor{red}{$\downarrow$ 0.3}\end{tabular}   & \begin{tabular}[c]{@{}c@{}}17.58G\\ \textcolor{gray}{$\rightarrow$ 0.0}\end{tabular} & \begin{tabular}[c]{@{}c@{}}1.94M\\ \textcolor{gray}{$\rightarrow$ 0.0}\end{tabular} \\ \bottomrule
\end{tabular}%
% }
\end{table*}

% \definecolor{gray3}{gray}{0.8}
\begin{table}[h!]
\centering
\setlength{\tabcolsep}{19pt}
\caption{Comparison of segmentation performance across single-task and multi-task models. The best and second best results are marked in \textbf{bold} and \underline{underline} respectively.}
\label{comparemore}
\begin{tabular}{lccc}
\toprule
\multirow{2}{*}{\textbf{Model}} & \textbf{Drivable Area}        & \textbf{Lane}                & \multirow{2}{*}{\textbf{Params} $\downarrow$} \\ \cmidrule{2-2}\cmidrule(l){3-3}
                                & \textbf{mIoU (\%)} $\uparrow$ & \textbf{IoU (\%)} $\uparrow$ &                                               \\ \hline
\multicolumn{4}{l}{\textit{Drivable Area Segmentation (Only)}} \\ 
PSPNet \cite{pspnet} & 89.6 & \xmark & -- \\
MultiNet \cite{multinet} & 71.6 & \xmark & -- \\
R-CNNP\(_\text{DA-Seg}\) \cite{yolop} & 90.2 & \xmark & -- \\
YOLOP\(_\text{DA-Seg}\) \cite{yolop} & 92.0 & \xmark & -- \\
YOLOv8\(_\text{segda}\) \cite{yoloveight} & 78.1 & \xmark & \textcolor{white}{0}3.26M  \\ 
YOLO-D \cite{yolold} & 92.4 & \xmark & -- \\ \hline
\multicolumn{4}{l}{\textit{Lane Segmentation (Only)}} \\ 
ENet \cite{enet} & \xmark & 14.6 & -- \\
ENET-SAD \cite{enetsad} & \xmark & 16.0 & -- \\
SCNN \cite{scnn} & \xmark & 15.8 & -- \\
YOLOv8\(_{segll}\) \cite{yoloveight} & \xmark & 22.9 & \textcolor{white}{0}3.26M  \\ 
YOLO-L \cite{yolold} & \xmark & 28.1 & -- \\ \hline
\multicolumn{4}{l}{\textit{Drivable Area and Lane Segmentation}} \\
DeepLabV3+ \cite{deeplab} & 90.9 & 29.8 & \textcolor{white}{0}15.4M \\
SegFormer \cite{segformer} & 92.3 & 31.7 & \textcolor{white}{00}7.2M \\
R-CNNP \cite{yolop} & 90.2 & 24.0 & -- \\
YOLOP\(_\text{Seg Only}\) \cite{yolop} & 91.6 & 26.5 & \textcolor{white}{0}5.53M \\
IALaneNet\(_{ConvNeXt-small}\) \cite{interactive} & 91.7 & \underline{32.5} & \textcolor{white}{0}39.9M \\
YOLOv8\(_\text{Multi}\) \cite{yoloveight} & 84.2 & 24.3 & -- \\
Sparse U-PDP\(_\text{w/o Detection}\) \cite{sparse} & 91.5 & 31.2 & -- \\
TwinLiteNet \cite{twin} & 91.3 & 31.1 & \textcolor{white}{0}0.44M \\ \hline
\multicolumn{4}{l}{\textit{Drivable Area and Lane Segmentation + Vehicle Detection}} \\ 
HybridNets \cite{hybridnets} & 90.5 & 31.6 & \textcolor{white}{0}13.8M \\
YOLOP \cite{yolop} & 91.5 & 26.2 & \textcolor{white}{00}7.9M \\
DRMNet \cite{drmnet} & 92.2 & 27.0 & \textcolor{white}{0}8.09M \\
YOLOPv2 \cite{yolopv2} & \textbf{93.2} & 27.3 & \textcolor{white}{0}38.9M \\
A-YOLOM(n) \cite{yolom} & 90.5 & 28.2 & \textcolor{white}{0}4.43M \\
A-YOLOM(s) \cite{yolom} & 91.0 & 28.8 & 13.61M \\
YOLOPX \cite{yolopx} & \textbf{93.2} & 27.2 & \textcolor{white}{0}32.9M \\
CenterPNets \cite{centerpnets} & 92.8 & 32.1 & \textcolor{white}{0}28.6M \\
Sparse U-PDP \cite{sparse} & \underline{92.9} & 32.4 & 12.05M \\ 
CFFM \cite{cffm} & 92.8 & 31.1 & - \\ 
TriLiteNet$_{base}$ \cite{trilitenet} & 92.4 & 29.8 & \textcolor{white}{0}2.35M \\ 
YOLOPv3 \cite{yolopv3} & \textbf{93.2} & 28.0 & \textcolor{white}{0}30.2M \\ 
\midrule
\rowcolor{gray3} TwinLiteNet$^+_{\text{Nano}}$ & 87.3 & 23.3 & \textbf{0.03M} \\
\rowcolor{gray3} TwinLiteNet$^+_{\text{Small}}$ & 90.6 & 29.3 & \underline{0.12M} \\
\rowcolor{gray3} TwinLiteNet$^+_{\text{Medium}}$ & 92.0 & 32.3 & 0.48M \\
\rowcolor{gray3} TwinLiteNet$^+_{\text{Large}}$ & \underline{92.9} & \textbf{34.2} & 1.94M \\
\bottomrule
\end{tabular}
\end{table}

\subsection{\textbf{More comparison}}
In Table~\ref{comparemore}, we present a comprehensive comparison of the proposed TwinLiteNet$^+$ model against a diverse set of representative methods selected based on task relevance. These include: (1) single-task models, which focus exclusively on either lane segmentation (e.g., \cite{enet,enetsad,scnn,yolom,yolold}) or drivable area segmentation (e.g., \cite{pspnet,multinet,yolop,yolom,yolold}); (2)  and models designed explicitly for drivable area and lane segmentation \cite{twin,interactive,lanenet,deeplab,segformer}, which share the same task formulation as our method; (3) multi-task models that jointly perform drivable area, lane, and object detection, such as \cite{hybridnets,yolop,yolopv2,yolopv3,yolopx,yolom,centerpnets,trilitenet,cffm,sparse}. This selection ensures that our evaluation captures related approaches, from task-specific to comprehensive multi-task perception models.

The results indicate that TwinLiteNet$^+$ achieves highly competitive performance, particularly within the lightweight model category. Notably, TwinLiteNet$^+_{\text{Nano}}$ and TwinLiteNet$^+_{\text{Small}}$ exhibit excellent trade-offs between model size and accuracy, with parameter counts of only 0.03M and 0.12M, respectively-making them well-suited for deployment on resource-constrained embedded systems. While single-task models are typically optimized for a specific objective, they lack the capacity to leverage complementary information across tasks, as discussed in Section~\ref{ablation_stvsmt}. In contrast, the multitask learning strategy employed in TwinLiteNet$^+$ enhances segmentation performance by enabling effective feature sharing, allowing it to outperform dedicated single-task counterparts. On the other hand, multitask models that incorporate object detection often require significantly more complex architectures and higher computational overhead. Thus, a multitasking design focused solely on segmentation tasks-such as TwinLiteNet$^+$-achieves a favorable balance between accuracy, efficiency, and architectural simplicity, making it more suitable for real-world applications on embedded platforms.

While TwinLiteNet$^+_{\text{Large}}$ achieves a strong mIoU of 92.9\% for drivable area segmentation, it falls slightly short of the 93.2\% achieved by high-capacity models such as YOLOPv2 \cite{yolopv2}, YOLOPv3 \cite{yolopv3}, and YOLOPX \cite{yolopx}. However, these models have a large number of parameters, exceeding 30 million, rendering them less practical for embedded deployment. Nevertheless, TwinLiteNet$^+_{\text{Large}}$ achieves the highest lane segmentation IoU in the entire benchmark (34.2\%), outperforming the second-best IALaneNet \cite{interactive} (32.5\%), despite IALaneNet requiring over 39.9M parameters-more than 20 times that of TwinLiteNet$^+{\text{Large}}$. 

These results provide a broader perspective by evaluating TwinLiteNet$^+$ against both single-task and multi-task approaches, including those that include object detection, further ensuring the effectiveness and scalability of our proposed lightweight model.

\section{Discussion and Conclusion}

\subsection{Limitations and future work}

Although TwinLiteNet$^+$ achieves a reasonable balance between accuracy and computational efficiency, there are still several limitations that should be addressed in future research. \textit{First}, the model is currently trained and evaluated under general environmental conditions and has not been optimized for specific scenarios. As a result, its performance tends to degrade in adverse weather conditions or cases with limited training samples. \textit{Second}, TwinLiteNet$^+$ currently focuses on only two perception tasks: drivable area segmentation and lane detection. While this design helps maintain a compact and efficient architecture, it lacks the comprehensiveness required by more advanced autonomous driving systems, which typically demand simultaneous handling of multiple tasks such as object detection, depth estimation, or traffic sign recognition. \textit{Third}, although the model is lightweight and efficient, it relies on manually designed architectural configurations. The lack of neural architecture search (NAS) techniques limits its adaptability to diverse deployment requirements in real-world settings.

Addressing these limitations will be essential further to enhance the model's performance and practical applicability. Doing so will not only improve TwinLiteNet$^+$ itself but also contribute to the advancement of more intelligent, efficient, and reliable multi-task perception systems for autonomous driving.

\subsection{Conclusion}

In this study, we presented TwinLiteNet$^+$, a novel and efficient multi-task segmentation model designed for drivable area and lane detection. By enhancing both the encoder and decoder architecture while maintaining a lightweight design, TwinLiteNet$^+$ achieves a strong balance between segmentation accuracy and computational efficiency. The model is offered in four scalable configurations, making it adaptable to a wide range of hardware-from high-performance GPUs to resource-constrained embedded systems. Extensive experiments on the BDD100K dataset demonstrate that TwinLiteNet$^+$ not only delivers competitive accuracy compared to existing methods but also ensures real-time performance suitable for practical deployment. These results highlight the model’s potential as a reliable perception module in intelligent driving systems, where both efficiency and accuracy are critical. This work contributes a promising solution for real-world autonomous driving applications, where streamlined architectures and performance-aware designs are increasingly essential.

\section{Acknowledgement}
This research was supported by The VNUHCM-University of Information Technology's Scientific Research Support Fund.

{\small
% \bibstyle{IEEEtran}
\bibliography{ref}
}
\end{document}